\documentclass{article}

\usepackage[final,main]{neurips_2026}
\usepackage[ruled,vlined,linesnumbered]{algorithm2e}

\usepackage{graphicx} %
\usepackage{notation}
\usepackage{etoolbox}
\usepackage{xparse} %
\usepackage{bbm}    %
\usepackage{tikz}
\usepackage{enumitem}
\usepackage{booktabs}
\usepackage{wrapfig}
\usepackage{amsthm}
\usepackage{amssymb}
\usepackage{natbib}
\usepackage{caption}
\usepackage{multirow}
\usepackage{array}
\usepackage{cancel}
\usepackage[american]{babel}
\usepackage{mathtools}
\usepackage[most]{tcolorbox}
\usepackage[dvipsnames]{xcolor}
\usepackage[backref=page,colorlinks=true,citecolor=blue]{hyperref}
\usepackage{cleveref}
\usepackage[title]{appendix}
\usepackage{minitoc}
\doparttoc %
\usepackage{listings}
\usepackage{longtable}
\usepackage{adjustbox}
\usepackage{csquotes}
\usepackage{makecell}
\usepackage[frozencache]{minted}
\setminted{fontsize=\small, linenos, breaklines}
\renewcommand{\mkbegdispquote}[2]{\itshape}

\tcbuselibrary{listingsutf8}

\newcommand{\revised}[1]{{#1}}

\crefname{figure}{Fig.}{Figs.}
\crefname{definition}{Def.}{Defs.}
\crefname{corollary}{Cor.}{Cors.}
\crefname{proposition}{Prop.}{Props.}
\crefname{observation}{Obs.}{Obs.}
\crefname{theorem}{Thm.}{Thms.}
\crefname{remark}{Remark}{Remarks}
\crefname{principle}{Principle}{Principles}
\crefname{lemma}{Lemma}{Lemmata}
\crefname{claim}{Claim}{Claims}
\crefname{table}{Tab.}{Tabs.}
\crefname{appendix}{App.}{App.}
\crefname{section}{Sec.}{Sec.}
\crefname{subsection}{Sec.}{Sec.}
\crefname{subsubsection}{Sec.}{Sec.}
\crefname{assumption}{Assumption}{Assumptions}
\crefname{algorithm}{Alg.}{Algs.}
\crefname{equation}{}{}
\crefname{enumi}{Step}{Steps}
\crefname{example}{Example}{Examples}

\newtheorem{proposition}{Proposition}[section]
\newtheorem{theorem}{Theorem}[section]

\newtheorem{corollary}{Corollary}[section]

\newcommand{\ourmethod}{PENEX}

\definecolor{myblue}{RGB}{30, 144, 255}
\newcommand{\nicecomment}[1]{\hfill\textcolor{myblue}{\texttt{// #1}} \\}
\newtcolorbox{centralequation}[1][]{
  colback=myblue!5,
  rounded corners,
  boxrule=0.5pt,
  top=0pt,
  bottom=4pt,
  #1 %
}

\newtcolorbox{codebox}{
  enhanced,
  colback=myblue!5,
  boxrule=0.4pt, arc=2mm,
  left=1mm, right=1mm, top=0.6mm, bottom=0.6mm,
  before skip=6pt, after skip=6pt
}

\newtcolorbox{remarkbox}[1][]{
  colback=white,
  rounded corners,
  boxrule=0.5pt
  #1 %
}

\newcolumntype{C}[1]{>{\centering\arraybackslash}p{#1}}

\title{PENEX: AdaBoost-Inspired Neural Network Regularization}

\author{
  Klaus-Rudolf Kladny\textsuperscript{1,2}
  \And
  Bernhard Schölkopf\textsuperscript{1,2,3}
  \And
  Michael Muehlebach\textsuperscript{1,4} \\ \\
  \textsuperscript{1} Max Planck Institute for Intelligent Systems, Tübingen \\
  \textsuperscript{2} Tübingen AI Center \\
  \textsuperscript{3} ELLIS Institute Tübingen \\
  \textsuperscript{4} Vienna University of Economics and Business\\
  \texttt{\{kkladny,bs,michaelm\}@tue.mpg.de}
}
\begin{document}

\renewcommand \partname{}
\renewcommand \thepart{}

\maketitle

\begin{abstract}
    AdaBoost sequentially fits so-called weak learners to minimize an exponential loss, which penalizes misclassified data points more severely than other loss functions like cross-entropy. Paradoxically, AdaBoost generalizes well in practice as the number of weak learners grows. In the present work, we introduce \textit{\underline{Pen}alized \underline{Ex}ponential Loss} (PENEX), a new formulation of the multi-class exponential loss that is theoretically grounded and, in contrast to the existing formulation, amenable to optimization via first-order methods, making it a practical objective for training neural networks. We demonstrate that PENEX effectively increases margins of data points, which can be translated into a generalization bound. Empirically, across computer vision and language tasks, PENEX improves neural network generalization in \emph{low-data regimes}, {matching and in some settings outperforming established regularizers at comparable computational cost}. {Our results highlight the potential of the exponential loss beyond its application in AdaBoost.}
\end{abstract}

\section{Introduction} \label{sec:introduction}

Regularization techniques improve generalization, typically at the cost of reduced in-sample performance~\citep{bengio2017deep}. There exists a variety of regularizers such as $\ell_2$ regularization~\citep{tikhonov1943stability, foster1961application}, early stopping~(e.g.,~\citet{bishop1995regularization}), RKHS norm penalization (e.g.,~\citet{SchSmo02}), adversarial training~\citep{szegedy2013intriguing}, dropout~\citep{srivastava2014dropout} and entropy regularization~\citep{meister2020generalized}. 

\begin{figure*}[t]
    \centering
    \includegraphics[width=14cm]{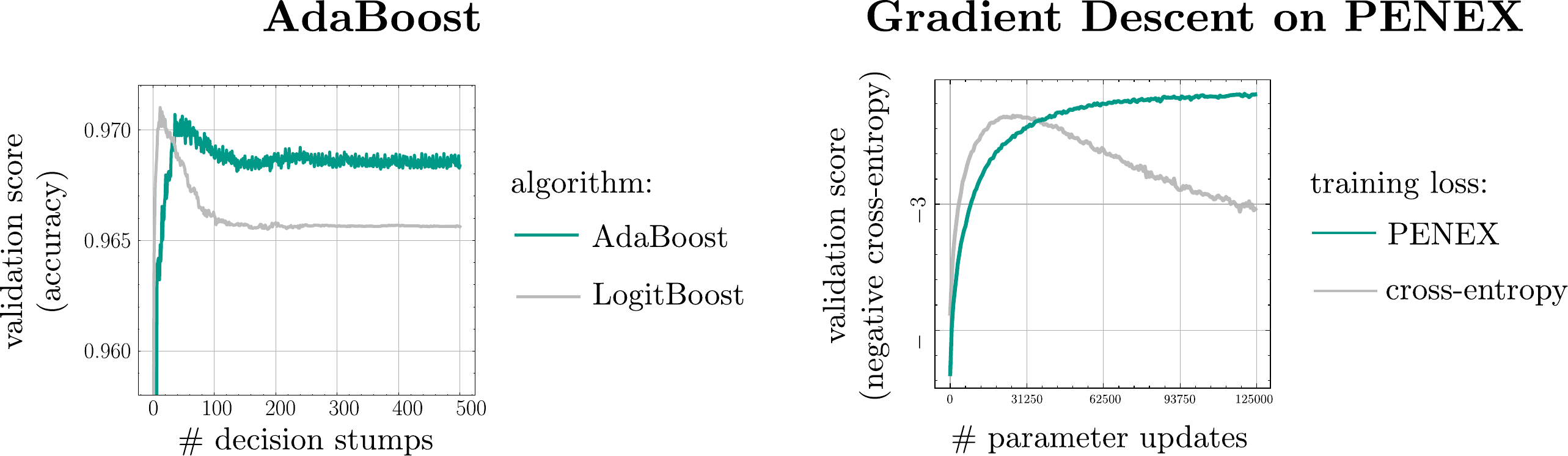}
    \caption{\textbf{AdaBoost vs. Gradient Descent on PENEX.} AdaBoost (left) builds a strong learner by sequentially fitting weak learners such as decision stumps and linearly combining them. Prior work has shown that this algorithm tends to be resilient to ``overfitting'', meaning that validation performance does not worsen as many decision stumps are added. \ourmethod{} (right) extends this phenomenon to deep neural network training by re-formulating the exponential loss into a formulation that is amenable to optimization via first-order methods. Just as for AdaBoost, the neural network does not overfit and keeps generalizing better in contrast to using cross-entropy loss.}
    \label{fig:main_visualization}
\end{figure*}

\begin{figure*}[t]
    \centering
    \includegraphics[width=\textwidth]{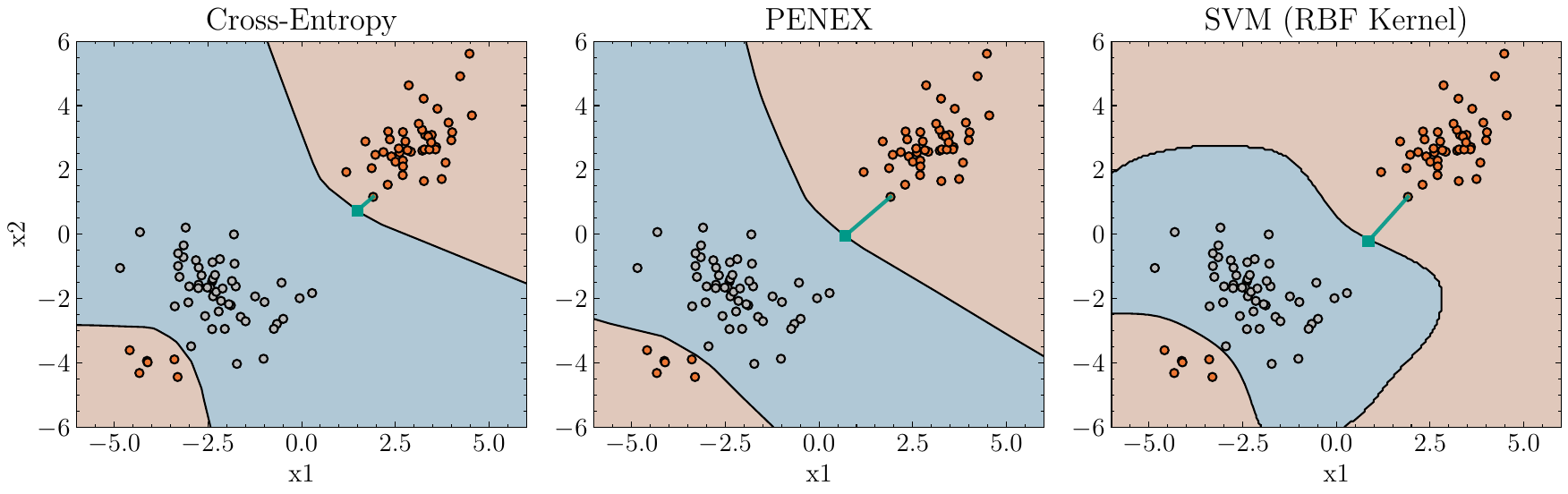}  %
    \caption{\textbf{Comparison of Margins.} Neural networks trained with \ourmethod{} (center) tend to implicitly increase the smallest margins (here, geometric margins, indicated for example point in green), in a similar way to support vector machines (right), here trained with a RBF kernel~\citep[p.~46]{SchSmo02}. Training via cross-entropy loss (left), in contrast, typically leads to smaller margins.}
    \label{fig:margins}
\end{figure*}

A seemingly unrelated branch of literature to regularization is boosting~\citep{schapire1990strength}: The method of generating a \textit{strong learner} by sequentially fitting \textit{weak learners}. A weak learner can be thought of as an inaccurate ``rule-of-thumb'' that barely does better than random guessing. A prime example of a weak learner is a single-split decision tree (``decision stump''), which classifies based on a single attribute split (e.g.,~\citet[p.~362]{hastie2017elements}; left part of~\cref{fig:main_visualization}). The goal of boosting is to train weak learners in order to gradually refine the predictions arising from a linear combination over all weak learners that have been fit so far (resulting in a single, strong learner). One of the best-known boosting algorithms for classification is AdaBoost~\citep{freund1995desicion}, which has been called the ``best off-the-shelf classifier in the world'' by Leo Breiman~\citep{friedman2000additive}. Although originally derived from the probably approximately correct learning framework~\citep{valiant1984theory},~\citet{breiman1999prediction} famously gave AdaBoost an interpretation in the language of empirical risk minimization~\citep{vapnik2013nature} by showing that AdaBoost minimizes an exponential loss function. This loss has been criticized for its lack of robustness with respect to outliers and misclassified data points~\citep[p.~662]{bishop2006pattern}. Counterintuitively, empirical studies have demonstrated that AdaBoost resists ``overfitting'' in practice as the number of weak learners increases~\citep{drucker1995boosting, breiman1996arcing, friedman2000additive}, thereby seemingly defying the Occam's razor principle by which simpler models should be favored over complex ones~(e.g.,~\citet[p. 343]{mackay2003information}). To resolve the paradox, various authors have developed theoretical arguments, {most of them based on margin properties~\citep{bartlett1998boosting, ratsch2001soft, rosset2003margin, grove1998boosting, onoda1998asymptotic, rosset2004boosting}:\footnote{Though alternative theories based on bias-variance theory~\citep{breiman1996arcing, breiman1996bias, bauer1999empirical} and self-averaging properties~\citep{wyner2017explaining} exist.} The margin is the minimum distance of a data point to any decision boundary. Due to the large penalty on miss-classifications, the model concentrates more on hard-to-learn examples, which play a similar role to support vectors~(\citet{cortes1995support},~\cref{fig:margins}). Thus, the tail of the margin distribution is lowered. A light tail of the margin distribution, in turn, guarantees good generalization~\citep[Thm. 9.2]{mohri2018foundations}.}

{Motivated by the works above}, we take the view that AdaBoost does not generalize well \textit{in spite} of the exponential loss, but \textit{because} of it, in the sense that it acts as a regularizer. The main contribution of the present work is to effectively translate this regularizing quality to training deep neural network classifiers. Specifically, we introduce the \textit{Penalized Exponential Loss} (\ourmethod{}), which reformulates the objective of the multi-class exponential loss~\citep{hastie2009multi} to a neural network-friendly variant by replacing a hard sum constraint by a SumExp penalty. We show that \ourmethod{} is self-calibrating, meaning that at the population level its optimal logits can be rescaled in closed form to recover the true class-probability distribution, and that it is Fisher consistent~\citep{fisher1922mathematical}, i.e., its optimal logits induce the Bayes-optimal classifier~\citep{lin2004note, hastie2009multi}. We also prove that \ourmethod{} {bounds the tail of the margin distribution}, thereby providing a theoretical foundation for \ourmethod{}'s generalization ability.\footnote{We include a broader discussion about how this result differs from the known result by~\citet{bartlett1998boosting} in~\cref{appx:margin_result_difference}.} We propose to minimize \ourmethod{} via gradient descent, which, at first sight, appears to contrast the idea of sequentially fitting weak learners in the boosting framework. However, we show in~\cref{appx:rel_to_adaboost} that small gradient-based parameter increments implicitly parameterize a form of weak learner. This result draws an analogy in that training on \ourmethod{} keeps improving on generalization as the model is trained for many epochs, just as how AdaBoost keeps improving on generalization as many decision stumps are added. 

Empirically, we observe that~\ourmethod{} performs considerably better than the standard formulation of the exponential loss~\citep{hastie2009multi}. In low data regimes, \ourmethod{} often improves upon the performance of popular regularizers such as label smoothing~\citep{szegedy2016rethinking} and confidence penalty~\citep{pereyra2017regularizing} across computer vision and language modeling tasks.

We summarize our main contributions as follows:

\begin{itemize}[leftmargin=*]
        \item We introduce \ourmethod{}~(\cref{sec:PEL}), an unconstrained alternative to the optimization problem proposed by \citet{hastie2009multi} as stated in~\cref{eq:L_CEL} below. We demonstrate favorable theoretical properties of \ourmethod{}, specifically self-calibration~(\cref{prop:self_calibration}), Fisher consistency~(\cref{cor:fisher_consistency}) and its margin-promoting quality~(\cref{theorem:margin_max}). {The latter can be combined with an existing result by~\citet{mohri2018foundations} to provide a generalization bound~(\cref{cor:generalization_bound})}.
        \item We show that the optimal penalty parameter can be inferred from~\cref{theorem:margin_max}, in closed form. This insight motivates an online algorithm for estimating the optimal penalty parameter~(\cref{alg:penex_training}) and therefore eliminates one tuning parameter~(\cref{sec:theoretical_properties},~\cref{sec:practical_implementation}).
        \item We empirically compare the regularizing effect of \ourmethod{} with the classical formulation of the exponential loss and other regularizers, on diverse tasks ranging from computer vision to fine-tuning of a language model. Our results show that \ourmethod{} greatly outperforms the classical formulation of the exponential loss and often improves upon existing regularization techniques with similar computational cost, in low data regimes~(\cref{sec:experiments}). 
        \item Our empirical results effectively translate the well-known observation that AdaBoost generalizes well with increasing numbers of weak learners to deep learning.
\end{itemize}

\paragraph{Overview.}~\cref{sec:PEL} introduces our main contribution, the \ourmethod{} loss function. We demonstrate desirable properties of \ourmethod{} in~\cref{sec:theoretical_properties} and practical implementation in~\cref{sec:practical_implementation}. In~\cref{sec:related_work}, we provide an overview of related works. This section is followed by experiments in~\cref{sec:experiments} that demonstrate the advantages of our approach on computer vision and language models, in comparison to other regularizers and the existing formulation of the exponential loss~(\cref{sec:ablation_experiments}). Finally, we discuss limitations and future work in~\cref{sec:discussion} and conclude in~\cref{sec:conclusion}.

\section{Penalized Exponential Loss (PENEX)} \label{sec:PEL}

For {class logits} $f : \mathcal{X} \mapsto \RR^K$ mapping from input space $\Xcal$ to $K$ classes, the exponential loss (EX) is given as
\begin{equation} \label{eq:EL}
    \el \left( f; \, \alpha \right) \; \coloneqq \; \EEemp \left[ \exp \left\{ - \alpha f^{(y)}(\xb) \right\} \right],
\end{equation}
where $\EEemp [ z ] = n^{-1}\sum_{i=1}^n z_i$ denotes the empirical mean over a given labeled training set $\{ (\xb_i, y_i) \}_{i=1}^n$ and $\alpha > 0$ controls sensitivity. A multi-class version of the exponential loss, which we call linearly-constrained exponential loss (CONEX), has been derived by~\citet{hastie2009multi}:
\begin{equation} \label{eq:L_CEL}
\el \big( f; \, (K-1)^{-1} \big) \quad \text{subject to} \quad \sum_{j=1}^K f^{(j)}(\xb) = 0, \quad \forall \xb \in \Xcal.
\end{equation}
The intuition behind the constraint in~\cref{eq:L_CEL} is that it acts as a barrier preventing the $f^{(j)}$ from diverging without bounds: {without the constraint, the exponential loss~\cref{eq:EL} can be decreased indefinitely by sending all logits to infinity}. In the present work, we propose an alternative to CONEX, which we refer to as penalized exponential loss (\ourmethod{}). It penalizes EX~\cref{eq:EL} by a SumExp $\sumexp(\xb) \coloneqq \sum_{j=1}^K \exp\{f^{(j)}(\xb)\}$ over $f^{(j)}$, specifically
\begin{centralequation}
    \begin{equation} \label{eq:PEL}
        \pel(f; \, \alpha, \rho) \coloneqq \el(f; \, \alpha) + \rho \EEemp \Big[ \sum_{j=1}^K \exp\{f^{(j)}(\xb)\} \Big],
    \end{equation}
\end{centralequation}
where $\rho > 0$ is a penalty parameter. The penalty in~\cref{eq:PEL} fulfills a similar purpose as the zero-sum constraint in~\cref{eq:L_CEL}, but acts differently: PENEX penalizes individual logits for being large rather than requiring logits to ``cancel each other out''. The penalty is an alternative means of making the exponential loss~\cref{eq:EL} well-posed. {Therefore, the penalty in~\cref{eq:PEL} is not an optional ``regularization term'' as commonly used in methods such as weight decay, but rather acts as a barrier term preventing divergence of the optimization.} The main benefit of \ourmethod{}~\cref{eq:PEL} over CONEX~\cref{eq:L_CEL} is that \ourmethod{} avoids constraints, which would require more elaborate algorithms like the augmented Lagrangian method~\citep{hestenes1969multiplier}. Instead,~\cref{eq:PEL} poses just an objective function, which can be optimized directly using gradient-based algorithms like Adam~\citep{kingma2014adam}.

\subsection{Theoretical Properties} \label{sec:theoretical_properties}
\begin{wrapfigure}{r}{0.49\textwidth}
    \vspace{-10pt}
    \centering
    \includegraphics[width=0.99\linewidth]{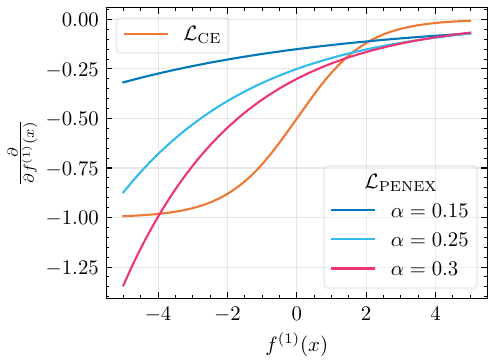}
    \caption{\textbf{CE vs. \ourmethod{}.} We consider the binary case ($K=2$) with $f^{(2)}(x) \equiv 0$, for a single $x$ and $y=1$. \ourmethod{} penalizes errors more than cross-entropy.}
    \label{fig:penex_deriv}
    \vspace{-5pt}
\end{wrapfigure}
The following proposition provides a theoretical ground for~\ourmethod{}:
\begin{proposition}[Self-Calibration] \label{prop:self_calibration}
For any \revised{fixed} $\rho>0$ and $\alpha>0$, and for any $\xb$ in the support of the distribution of $\xb$, the minimizer of the population equivalent of~\cref{eq:PEL},
called $f_*$:
\begin{equation} \label{eq:pop_minimizer}
    f_* \coloneqq \arg \min_f \, \EE [\pel(f; \, \alpha, \rho)],
\end{equation}
is unique and recovers the true class probabilities $P(y \mid \xb)$ via
\begin{equation*}
    P(y \mid \xb)\;\propto\; \exp\!\left\{(1+\alpha)\, f_*^{(y)}(\xb)\right\},
\end{equation*}
i.e., the $\fopt^{(y)}(\xb)$ recover the log-posteriors at temperature $(1 + \alpha )^{-1}$, up to a constant shift.
\end{proposition}
A proof for~\cref{prop:self_calibration} is supplied in~\cref{appx:fisher_consistency}. \cref{prop:self_calibration} shows that \ourmethod{} provides a closed-form temperature re-scaling to recover the true class probabilities, on the population level. In addition,~\cref{prop:self_calibration} lays the ground for showing that \ourmethod{} is Fisher consistent:
\begin{corollary}[Fisher Consistency] \label{cor:fisher_consistency}
Under the assumptions of \cref{prop:self_calibration}, PENEX~\cref{eq:PEL} is \textit{Fisher consistent}
according to the definition of~\citet{hastie2009multi}, i.e., $f_*$~\cref{eq:pop_minimizer}
induces the Bayes-optimal classifier via
\begin{equation*}
    \arg\max_i P(i\mid \xb) \; = \; \arg\max_i f_*^{(i)}(\xb).
\end{equation*}
\end{corollary}
\cref{cor:fisher_consistency} follows directly from \cref{prop:self_calibration} since $(1+\alpha)>0$ and so
\begin{equation*}
\arg\max_i f_*^{(i)}(\xb) \quad
= \quad \arg\max_i \, \exp \bigl\{(1+\alpha)\,f_*^{(i)}(\xb)\bigr\}
\quad = \quad \arg\max_i P(i \mid \xb)\, .
\end{equation*}
While other loss functions such as cross-entropy $\cel(f) \coloneqq -\EEemp \big[ \log \big( \Phat( y \, | \, \xb ) \big) \big]$ with $\Phat( y \, | \, \xb ) \propto \exp \{ f^{(y)}(\xb) \}$ are Fisher consistent as well, the solutions on finite datasets usually differ, as can be inferred from~\cref{fig:penex_deriv}: When the ``correct'' logit grows negative ($f^{(1)}(\xb) \rightarrow -\infty$), the gradient of \ourmethod{} keeps decreasing. The cross‐entropy gradient, however, levels out at $-1$.

A key aspect in the generalization ability of \ourmethod{} lies in a margin-emphasizing quality. In the present work, we define the margin $m(\xb, y)$ of an example $(\xb, y)$ as the minimum difference of the true logit to any ``false'' logit,
\begin{equation*}
    m_f(\xb, y) \; \coloneqq \; f^{(y)}(\xb) 
    \, - 
    \, \max_{j \neq y} f^{(j)}(\xb).
\end{equation*}
The next theorem shows that \ourmethod{} effectively aims at avoiding small margins:

\begin{theorem}[{Margin Tail Bound}] \label{theorem:margin_max}
    For any $\gamma > 0$, it holds that
    \begin{equation} \label{eq:margin_bound}
        \PPast (m_{f}(\xb, y) \leq \gamma) \leq e^{\gamma \frac{\alpha}{\alpha + 1}} \rho^{-\frac{\alpha}{\alpha + 1}} \pel(f; \, \alpha, \rho),
    \end{equation}
    where $\PPast$ denotes empirical probability.
\end{theorem}
A proof is provided in~\cref{appx:margin_max}. Intuitively,~\cref{theorem:margin_max} says that the fraction of data points with a small margin is low, if \ourmethod{} has a small value for a model $f$. {Defining the expected miss-classification error $\Rcal$ as
\begin{equation} \label{eq:risk}
    \Rcal(f) \; \coloneqq \; \EE [\indfonearg{\mathrm{arg} \, \mathrm{max}_i \, f^{(i)}(\xb) \, \neq \, y}], 
\end{equation}
we can combine~\cref{theorem:margin_max} with a known result from statistical learning~\citep[Thm. 9.2]{mohri2018foundations} to obtain the following generalization bound:
\begin{corollary}[Generalization Bound] \label{cor:generalization_bound}
    Assume $f \in \Fcal$ and denote Rademacher complexity of $\Fcal$ by $\mathfrak{R}$. Then, for any $\delta \in (0; 1), \rho > 0,\gamma > 0$, it holds with probability at least $1-\delta$ that
    \begin{equation*}
        \Rcal (f) \; \leq \; e^{\gamma  \frac{\alpha}{\alpha + 1}} \rho^{-\frac{\alpha}{\alpha + 1}} \pel(f; \, \alpha, \rho) \, + \, 
        \frac{4K}{\gamma} \mathfrak{R}(\Fcal) \, + \, \sqrt{\frac{\log(1/\delta)}{2n}}. 
    \end{equation*}
\end{corollary}
}

{We interpret \cref{cor:generalization_bound} by minimizing the right-hand side over $\gamma>0$. The choice of $\gamma$ mediates a trade-off: the capacity term $\frac{4K}{\gamma}\mathfrak{R}(\Fcal)$ decreases as $1/\gamma$, whereas the leading term increases as $\exp( \gamma \tfrac{\alpha}{\alpha+1} )$. Consequently, the bound suggests an intermediate margin threshold $\gamma_\ast$. Notably, achieving a smaller PENEX value decreases the prefactor of the exponential term, which shifts $\gamma_\ast$ toward larger values and thereby reduces the capacity term $\frac{4K}{\gamma_\ast}\mathfrak{R}(\Fcal)$.} 

In addition to providing a generalization bound, \cref{theorem:margin_max} suggests a principled choice of $\rho$:
\begin{proposition}[Optimal Penalty Parameter] \label{prop:optimal_penalty}
    The parameter $\rho$ that minimizes the upper bound in~\cref{eq:margin_bound} is
    \begin{equation} \label{eq:optimal_penalty}
        \rho_*(\alpha, f) \; = \; \alpha \frac{\EEemp[\el(f; \alpha)]}{\EEemp[\sumexp(f; \xb)]}.
    \end{equation}
\end{proposition}
This result follows from realizing that the upper bound in~\cref{eq:margin_bound} is convex and differentiable in $\rho$, thus we can solve for $\rho_*$ by setting its derivative to zero.~\cref{prop:optimal_penalty} has a practically relevant consequence in that it avoids the necessity of tuning $\rho$, as we demonstrate in the following section~(\cref{sec:practical_implementation}).

\subsection{Practical Implementation} \label{sec:practical_implementation}

\cref{prop:optimal_penalty} motivates dynamically estimating $\rho_*$ during training. In practice, however, the optimal penalty parameter in~\cref{eq:optimal_penalty} is expensive to compute directly, because $f$ changes during training and so $\rho_*$ would have to be re-computed on the entire data set after each parameter update. Thus, we rely on online batch statistics for estimation. To reduce variance, we use exponential moving averaging (EMA) and clipping, as shown in~\cref{alg:penex_training}. The weights for EMA and clipping are fixed for all experiments, i.e., we do not treat them as tuning parameters. We note that~\cref{alg:penex_training} can be implemented in a stateful loss object and \textit{does not require a custom training loop}. We refer to~\cref{appx:penex_practical} for more details regarding the implementation.

\begin{algorithm}[t]
\DontPrintSemicolon
\caption{PENEX Training}
\label{alg:penex_training}

\KwIn{
EMA factor $\beta > 0$; sensitivity $\alpha > 0$; 
clipping parameters $0 < \rho_{\min} < \rho_{\max}$; 
horizon $T \in \NN^+$; initialization $\theta_0$; optimizer $\texttt{optim}$; 
training set $\Dcal$; $\epsilon > 0$;
}

\BlankLine

\For{\(t \leftarrow 1, 2, \ldots, T\)}{
    $\Dcal_{\mathrm{b}} \overset{i.i.d.}{\sim} \Dcal$
    \nicecomment{sample mini-batch}

    $\displaystyle
    \rho' \gets
    \alpha
    \frac{
        \el^{\Dcal_{\mathrm{b}}}(f; \alpha)
    }{
        \EEemp_{\Dcal_{\mathrm{b}}}[\sumexp(f; \mathbf{x})] + \epsilon
    }$
    \nicecomment{\cref{eq:optimal_penalty}}

    \If{$t = 1$}{
        $\rho_0 \gets \rho'$ \;
    }

    $\rho'_t \gets (1 - \beta)\rho_{t-1} + \beta\rho'$
    \nicecomment{EMA}

    $\rho_t \gets
    \min \{
        \max \{ \rho'_t, \rho_{\min} \},
        \rho_{\max}
    \}$
    \nicecomment{clip}

    \BlankLine

    $\theta_t \gets
    \optim \left(
        \pel^{\Dcal_{\mathrm{b}}}
        (\theta_{t-1}; \alpha, \rho_t)
    \right)$ \;
}

\BlankLine
\Return $\theta_T$
\end{algorithm}

Also, the practical implementation implements the fixed temperature-rescaling suggested by~\cref{prop:self_calibration}. That is, during inference, estimates of class logits are always multiplied by $1 + \alpha$ before feeding them into the softmax function (more details are provided in~\cref{appx:penex_inference}).

\section{Related Work} \label{sec:related_work}

The exponential loss is strongly associated with AdaBoost. Thus, besides discussing prior works on gradient-based exponential loss minimization and margin maximization, we cover related works on the intersection of boosting and deep learning. 

\paragraph{Exponential Loss and Gradient Descent.} Prior works have highlighted a relationship between AdaBoost and gradient descent. Specifically,~\citet{mason1999boosting} discovered that binary AdaBoost can be thought of as an approximate gradient descent on the exponential loss, where the weak learners minimize an inner product with the gradient at the specific step.~\citet{friedman2001greedy} has made a similar observation and subsequently generalized this method to general loss functions, now commonly known as gradient boosting. Our work takes the reverse direction that gradient descent can also be interpreted as a form of boosting. Another line work by~\citet{soudry2018implicit, ji2018gradient, gunasekar2018characterizing, nacson2019convergence, cai2025implicit} developed theoretical results based on margin theory for performing gradient descent on the binary exponential loss~\cref{eq:EL} with linear/near-homogeneous models. This contrasts the focus of the present work, which specifically considers multi-class problems, without imposing assumptions on the underlying neural networks. Also, our ablation studies indicate that the existing formulation of the exponential loss studied in these listed works is not amenable to effective neural network training~(see~\cref{sec:ablation_experiments}).

{\paragraph{Margin Maximization in Deep Learning.} The idea of explicitly maximizing the classification margin in the context of deep learning has been explored in prior work: One line of work studies the effects of the spectral norm of the network’s Jacobian matrix~\citep{sokolic2017robust} or network depth~\citep{sun2016depth} on the classification margin to motivate regularization techniques.~\citet{elsayed2018large} introduce a large‑margin objective that enforces margins at selected intermediate layers. The method adds several hyperparameters and requires computing per‑example gradient norms with respect to those activations (scaling with the number of layers and top‑k classes), making the method computationally demanding on deeper architectures. Another line of work by~\citet{jiang2018predicting} shows that the distribution of the margin correlates with the generalization gap, which inspired~\citet{lyu2022improving} to develop a loss that penalizes deviations of the margin from its expected value. {While similarly lightweight to~\ourmethod{}, the objective of~\citet{lyu2022improving} does not recover the full multiclass probability distribution, thereby leaving the problem of calibration un-resolved~(see \cref{appx:margin_based} for experimental evidence).}}

\paragraph{Boosting in Deep Learning.}~\citet{tolstikhin2017adagan, grover2018boosted, giaquinto2020gradient} construct a ``strong'' generative model by greedily fitting multiple ``weak'' generative models in a boosting-like fashion. Similarly, but with classifiers instead of generative models, dense~\citep{schwenk1997adaboosting} and convolutional neural networks~\citep{drucker1992improving, lecun1995learning, taherkhani2020adaboost, sun2019adagcn} have been used as weak learners for Boosting. In the context of language models, boosting has further been applied to prompt generation~\citep{pitis2023boosted, zhang2024prefer}, text classification~\citep{hou2023promptboosting, huang2020boostingbert}, text-summary based prediction~\citep{manikandan2023language} and machine teaching~\citep{agrawal2024ensemw2s}. All of these studies differ from the present work in what is considered as a ``weak learner''. In the present work, this notion only arises implicitly~(\cref{appx:rel_to_adaboost}) and is given as the Jacobian, multiplied by a small parameter increment~\cref{eq:incremental_optimization_penex}, instead of existing approaches that consider explicitly combining entire models {(such as decision trees and neural networks)} to gradually build more complex ones.
\vspace{-0.5em}
\section{Experiments} \label{sec:experiments}

\begin{figure*}[t]
    \centering
    \includegraphics[width=\textwidth]{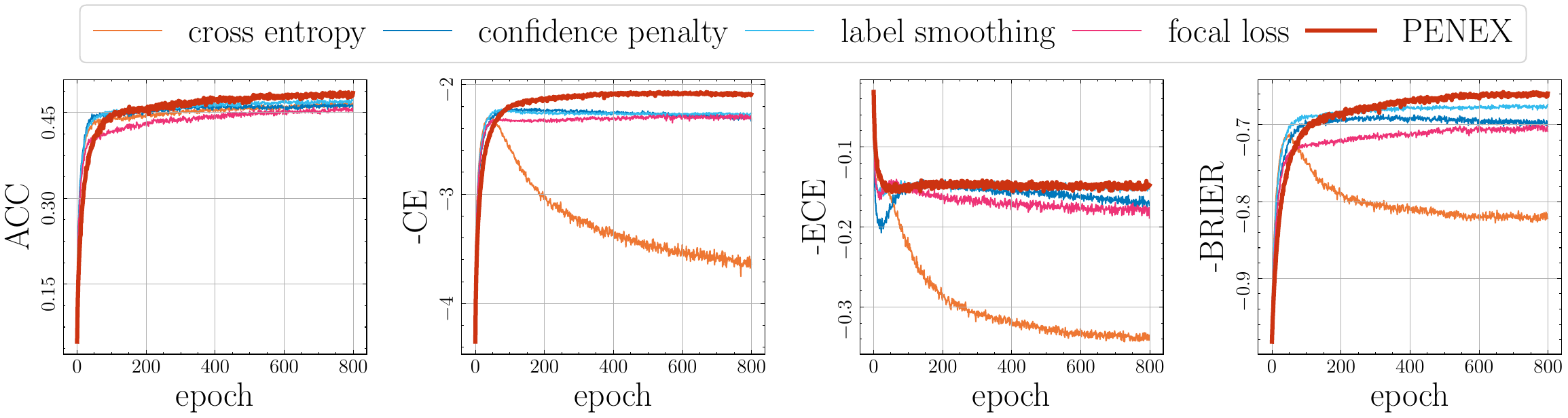}
    \caption{\textbf{Performance Analysis on CIFAR-100.} \underline{Larger means better}. Results are computed from validation data. All hyperparameters have been tuned individually. \ourmethod{} (thick red) is an effective regularizer with often better generalization than other common regularization techniques (thin), and shows no signs of ``overfitting'' like cross-entropy training (orange) }
    \label{fig:metric_curves}
\end{figure*}

We aim to answer the following questions: \textbf{1)} How does~\ourmethod{} compare with other regularization methods? \textbf{2)} How does optimizing PENEX~\cref{eq:PEL} compare to optimizing CONEX~\cref{eq:L_CEL} via constrained optimization? \textbf{3)} How does PENEX perform under label noise? \textbf{4)} How does \ourmethod{} perform for fine-tuning foundation models? \textbf{5)} How does the value of $\alpha$ affect training? \revised{\textbf{6)} How do other penalty formulations compare with the SumExp penalty term? \textbf{7)} How does \cref{alg:penex_training} compare with tuning a fixed $\rho$? \textbf{8)} How sensitive is \cref{alg:penex_training} with respect to EMA and clipping parameters?}

Experimental results addressing questions \textbf{5)}-\textbf{8)} are shown in~\cref{appx:parameter_analysis},~\cref{appx:different_penalties},~\cref{appx:fixed_rho} and~\cref{appx:ema_clip_ablation}, respectively. All other questions are answered in the main text.

\paragraph{Reproducibility.} All code can be found on GitHub at \href{https://github.com/rudolfwilliam/penex}{https://github.com/rudolfwilliam/penex}.

\vspace{-0.5em}
\subsection{Setup} \label{sec:experiments_setup}

In the main text, we provide a high-level description of the experimental setup. More details can be found in~\cref{appx:experimental_details}. 

\paragraph{Baselines.} {For fairness of comparison, we restrict ourselves to baselines that can be expressed as a single, architecture‑invariant loss function that does not require additional forward or backward passes (more details in~\cref{appx:runtime}).} \textbf{1)} Cross-entropy (CE). \textbf{2)} Confidence penalty~(\citet{pereyra2017regularizing}), \textbf{3)} Label smoothing~(\citet{szegedy2016rethinking}), and \textbf{4)} Focal loss~\citep{ross2017focal}. The latter three are well-established regularization techniques.
\vspace{-1em}
\paragraph{Tasks.} \textbf{1-2)} \underline{CIFAR-10/100}~\citep{krizhevsky2009learning}: Image classification with either $10$ or $100$ classes, using a convolutional neural network. An additional experiment with a residual neural network~\citep{he2016deep} is shown in~\cref{appx:alternative_architecture_experiment}. \textbf{3)} \underline{Noisy CIFAR-10}: CIFAR-10 with 10\% of training labels flipped uniformly at random. The test set does not have flipped labels. Further noise settings are covered in~\cref{appx:additional_noise}. \textbf{4)} \underline{Pathology Image Classification}~(PathMNIST;~\citet{kather2019predicting, medmnistv2}): Recognizing $9$ different tissue types from $10,000$ non-overlapping image patches from hematoxylin \& eosin stained histological images, using a convolutional neural network. \textbf{5)} \underline{ImageNet}~(\citet{5206848}): Recognizing $1000$ different classes from $\approx 14$ million images, using a vision transformer~\citep{Dosovitskiy2021image}. \textbf{6)} \underline{News Classification}~(BBC News;~\citet{greene2006practical}): Fine-tuning a pretrained RoBERTa-base language transformer encoder~\citep{liu2019roberta} for news article classification with $5$ classes.

\paragraph{Metrics.} \textbf{1)} \underline{Accuracy} (ACC): Proportion of correctly classified points {($1 - \hat{\Rcal}(f)$; see~\cref{eq:risk})}. \textbf{2)} \underline{Cross-Entropy} (CE): Corresponds to $\cel$, which (for classification) is equivalent to the \textit{negative log-likelihood}. \textbf{3)} \underline{Brier Score}~(BRIER;~\citet{brier1950verification}): Mean squared difference between predicted class probabilities and one-hot encoded labels. \textbf{4)} \underline{Expected Calibration Error}~(ECE;~\citet{guo2017calibration}): Measures the discrepancy between a model’s predicted confidence and its actual accuracy. \textit{We change the sign in some metrics, such that a larger value indicates better performance for all metrics.}

\paragraph{Hyperparameter Tuning.} Each method's hyperparameters (except for the learning rate, which we keep fixed for all methods) are tuned on the validation set using the tree-structured Parzen estimator (e.g.,~\citet{watanabe2023tree}). We minimize cross-entropy loss over a fixed amount of epochs. For \ourmethod{}, we always estimate $\rho^* (\alpha)$ according to~\cref{alg:penex_training} (see~\cref{appx:penex_practical} for details). Thus, \textit{\ourmethod{} only requires tuning the sensitivity $\alpha$}. Tuning is performed individually for each experiment and method, except for ImageNet, where we use default parameters for each method~(see~\cref{appx:imagenet_params}).

\subsection{Main Results} \label{sec:experiments_results}

\ourmethod{} often achieves regularization performance that outperforms other regularization techniques (though taking a bit longer to converge), as can be seen in~\cref{fig:metric_curves}, showing multiple validation curves for CIFAR-100~(analogous plots for all others datasets are shown in~\cref{appx:all_validation_curves}) as a function of the number of training epochs. We also provide test results in~\cref{tab:test_performance} that confirm the results from~\cref{fig:metric_curves} on multiple datasets, domains and metrics. Remarkably,~\ourmethod{} also outperforms other regularizers in the setting of artificial label noise. The only experiment where \ourmethod{} performs slightly worse than other methods is the ImageNet experiment. However, ImageNet is a large data set and other regularization methods do also not benefit generalization much in this setting.

\begin{table*}[t]
\centering
\caption{\textbf{Test Set Performance.} \underline{Larger means better, best is bold}. Mean test result $\pm 1$ standard deviation after $200$ training epochs (for ImageNet: $300$ epochs). Standard deviation is obtained by running $100$ bootstrap evaluations~\citep{tibshirani1993bootstrap}. Hyperparameters are tuned on validation data (except for ImageNet, see~\cref{appx:imagenet_params}).}
\vspace{0.3em}
\label{tab:test_performance}
\resizebox{0.99 \textwidth}{!}{
\begin{tabular}{ c || c || c | c | c | c | c | c }
\toprule
\textbf{Method} & \textbf{Metric} & \textbf{CIFAR-10} & \textbf{Noisy CIFAR-10} & \textbf{CIFAR-100} & \textbf{PathMNIST} & \textbf{ImageNet} & \textbf{BBC News} \\
\midrule
\multirow{4}{*}{CE}
 & ACC & $ 0.785 \pm 0.004$ & $ 0.724 \pm 0.004$ & $ 0.443 \pm 0.004$ & $ 0.826 \pm 0.004$ & $ 0.648 \pm 0.000$ & $ 0.967 \pm 0.007$ \\
 & -ECE & $ -0.162 \pm 0.003$ & $ -0.179 \pm 0.003$ & $ -0.287 \pm 0.003$ & $ -0.151 \pm 0.004$ & $ -0.200 \pm 0.000$ & $ \bf{-0.032} \pm 0.006$ \\
 & -CE & $ -1.004 \pm 0.024$ & $ -1.125 \pm 0.019$ & $ -3.072 \pm 0.034$ & $ -2.018 \pm 0.130$ & $ -2.175 \pm 0.000$ & $ -0.109 \pm 0.024$ \\
 & -BRIER & $ -0.346 \pm 0.006$ & $ -0.424 \pm 0.006$ & $ -0.794 \pm 0.006$ & $ -0.300 \pm 0.007$ & $ -0.536 \pm 0.000$ & $ -0.051 \pm 0.011$ \\
\midrule
\multirow{4}{*}{\makecell{label \\ smoothing}}
 & ACC & $ 0.789 \pm 0.004$ & $ 0.747 \pm 0.004$ & $ 0.451 \pm 0.005$ & $ 0.829 \pm 0.004$ & $ \bf{0.651} \pm 0.000$ & $ 0.970 \pm 0.006$ \\
 & -ECE & $ -0.112 \pm 0.002$ & $ -0.183 \pm 0.003$ & $ \bf{-0.147} \pm 0.002$ & $ -0.109 \pm 0.002$ & $ -0.159 \pm 0.000$ & $ -0.033 \pm 0.006$ \\
 & -CE & $ -0.657 \pm 0.011$ & $ -0.889 \pm 0.008$ & $ -2.292 \pm 0.019$ & $ \bf{-0.589} \pm 0.012$ & $ -2.180 \pm 0.000$ & $ -0.115 \pm 0.022$ \\
 & -BRIER & $ -0.300 \pm 0.005$ & $ -0.384 \pm 0.004$ & $ -0.692 \pm 0.004$ & $ -0.255 \pm 0.005$ & $ -0.537 \pm 0.000$ & $ -0.049 \pm 0.010$ \\
\midrule
\multirow{4}{*}{\makecell{confidence \\ penalty}}
 & ACC & $ 0.786 \pm 0.004$ & $ 0.733 \pm 0.004$ & $ 0.449 \pm 0.006$ & $ 0.828 \pm 0.004$ & $ 0.641 \pm 0.000$ & $ \bf{0.974} \pm 0.006$ \\
 & -ECE & $ -0.130 \pm 0.002$ & $ -0.149 \pm 0.003$ & $ -0.152 \pm 0.002$ & $ -0.110 \pm 0.003$ & $ -0.161 \pm 0.000$ & $ -0.050 \pm 0.005$ \\
 & -CE & $ -0.731 \pm 0.015$ & $ -0.866 \pm 0.009$ & $ -2.254 \pm 0.018$ & $ -0.917 \pm 0.047$ & $ -1.972 \pm 0.000$ & $ -0.094 \pm 0.015$ \\
 & -BRIER & $ -0.317 \pm 0.005$ & $ -0.385 \pm 0.004$ & $ -0.695 \pm 0.005$ & $ -0.262 \pm 0.005$ & $ -0.518 \pm 0.000$ & $ \bf{-0.042} \pm 0.008$ \\
\midrule
\multirow{4}{*}{\makecell{focal \\ loss}}
 & ACC & $ 0.778 \pm 0.004$ & $ 0.708 \pm 0.004$ & $ 0.428 \pm 0.005$ & $ 0.803 \pm 0.004$ & $ 0.647 \pm 0.000$ & $ 0.970 \pm 0.006$ \\
 & -ECE & $ -0.117 \pm 0.002$ & $ -0.165 \pm 0.003$ & $ -0.161 \pm 0.003$ & $ -0.112 \pm 0.003$ & $ \bf{-0.093} \pm 0.000$ & $ -0.051 \pm 0.005$ \\
 & -CE & $ -0.661 \pm 0.010$ & $ -0.905 \pm 0.008$ & $ -2.341 \pm 0.022$ & $ -0.939 \pm 0.050$ & $ \bf{-1.578} \pm 0.000$ & $ \bf{-0.092} \pm 0.014$ \\
 & -BRIER & $ -0.313 \pm 0.005$ & $ -0.423 \pm 0.004$ & $ -0.723 \pm 0.005$ & $ -0.291 \pm 0.006$ & $ \bf{-0.482} \pm 0.000$ & $ \bf{-0.042} \pm 0.008$ \\
\midrule
\multirow{4}{*}{\underline{PENEX}}
 & ACC & $ \bf{0.793} \pm 0.004$ & $ \bf{0.766} \pm 0.004$ & $ \bf{0.460} \pm 0.005$ & $ \bf{0.833} \pm 0.004$ & $ 0.590 \pm 0.000$ & $ 0.968 \pm 0.006$ \\
 & -ECE & $ \bf{-0.109} \pm 0.002$ & $ \bf{-0.131} \pm 0.002$ & $ \bf{-0.147} \pm 0.003$ & $ \bf{-0.100} \pm 0.003$ & $-0.107 \pm 0.000$ & $ -0.034 \pm 0.006$ \\
 & -CE & $ \bf{-0.646} \pm 0.012$ & $ \bf{-0.716} \pm 0.009$ & $ \bf{-2.140} \pm 0.018$ & $ -1.200 \pm 0.089$ & $ -1.853 \pm 0.000$ & $ -0.124 \pm 0.025$ \\
 & -BRIER & $ \bf{-0.299} \pm 0.005$ & $ \bf{-0.332} \pm 0.004$ & $ \bf{-0.685} \pm 0.004$ & $ \bf{-0.251} \pm 0.006$ & $-0.553 \pm 0.000$ & $ -0.055 \pm 0.011$ \\
\bottomrule
\end{tabular}
}
\end{table*}

\subsection{Ablation Studies} \label{sec:ablation_experiments}

We compare \ourmethod{} with the following ablations:
\textbf{1--3)} Optimizing CONEX~\cref{eq:L_CEL} via \textbf{1)} the augmented Lagrangian method~\citep{hestenes1969multiplier}; or \textbf{2)} the squared penalty method. \textbf{3)} Fixing $f^{(K)}(\xb) \equiv -\sum_{j=1}^{K-1}f^{(j)}(\xb)$. \textbf{4)} Optimizing EX~\cref{eq:EL} directly, neither with constraint nor penalty. Details for the implementation of each baseline are provided in~\cref{appx:baselines}.

\begin{figure}[t]
    \centering
    \includegraphics[width=0.99\textwidth]{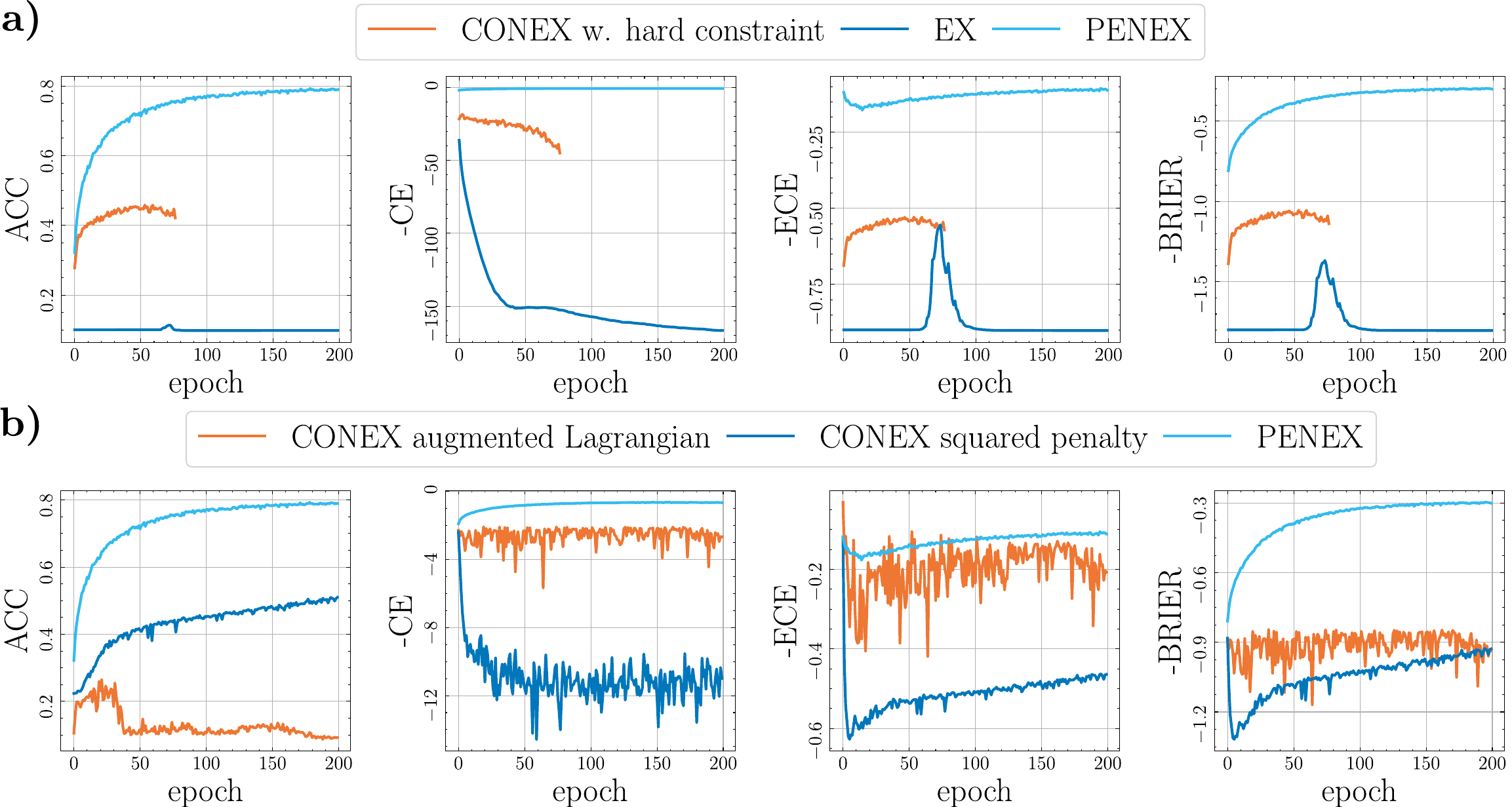}
    \caption{\textbf{Ablation Studies on CIFAR-10.} \underline{Larger means better}. \textbf{a)} Optimizing the exponential loss without any constraint or with the constraint encoded into the model architectures break down early on during training (no results are shown for the hard constraint method starting at around epoch $80$ because of \texttt{NaN} model outputs). \textbf{b)} While constrained optimization algorithms based on CONEX~\cref{eq:L_CEL} are trainable in principle, they reach worse optima and suffer from less stable training, in addition to requiring tuning more parameters~(see~\cref{appx:hparams}).}
    \label{fig:metric_curves_ablations}
\end{figure}

We perform all ablation studies for the CIFAR-10 dataset only. The results, displayed in~\cref{fig:metric_curves_ablations}, confirm the claim that \ourmethod{} is favorable over the CONEX formulation~\cref{eq:L_CEL} in the context of training deep neural networks. Enforcing the zero-sum constraint of CONEX in the model architecture (ablation \textbf{3)}) and directly minimizing the exponential loss~\cref{eq:EL} without any constraint (ablation \textbf{4)}) break down entirely: Using no penalty or constraint leads to all logits diverging without bounds, while the hard constraint method results in \texttt{NaN} logits around epoch $80$. Only the penalty and augmented Lagrangian approach for optimizing CONEX work at all, while still performing much worse than \ourmethod{}, in addition to requiring the tuning of at least one more hyperparameter~(see~\cref{appx:hparams}).

Overall, the ablation studies show that the \ourmethod{} formulation of the exponential loss is not merely a minor refinement over CONEX, but crucial for stable and effective neural network training.

\section{Discussion \& Limitations} \label{sec:discussion}

\paragraph{The Secret Sauce of AdaBoost.} Conventional wisdom treats AdaBoost’s exponential loss as a flaw, inspiring variants that swap the exponential for the logistic loss~\citep{friedman2000additive} and generalizations to arbitrary loss functions~\citep{mason1999boosting, friedman2001greedy}. Our evidence suggests the opposite.~\ourmethod{} typically outperforms standard cross-entropy training. Remarkably, the resulting accuracy gap is wider than the well-documented (and usually modest) difference between AdaBoost and LogitBoost, two algorithms that typically achieve similar performance (e.g.,~\citealp{friedman2000additive}). Explaining why the formulation introduced in the present work behaves so favorably in deep neural networks specifically remains an open direction for future work.

\paragraph{Label Noise.} High noise levels in the training labels have been regarded to be the Achilles' heel of AdaBoost~\citep{dietterich2000experimental, quinlan1996boosting, grove1998boosting, ratsch2001soft}, which does not translate to the present work: Neural networks trained with~\ourmethod{} handle artificial label noise even better than other regularizers (see~\cref{sec:experiments_results} and~\cref{appx:additional_noise}). This phenomenon may be related to the well-known ability of deep neural networks to fit random labels well~\citep{zhang2016understanding}. Further improving the resilience of~\ourmethod{} with respect to training label corruptions, for instance by incorporating ideas by~\citet{ratsch2001soft, freund1999adaptive}, may nevertheless be a promising direction for future work. 

\paragraph{Training Set Size.} The only experiment where \ourmethod{} does not perform better or equally well in comparison to other methods is the ImageNet experiment, indicating that \ourmethod{} is advantageous only in low-data regimes, such as CIFAR-100. We believe that developing remedies for this issue is another relevant direction for future research.

\paragraph{Beyond Classification.} We propose \ourmethod{} as a loss function for classification problems. Future work may explore adaptions of \ourmethod{} to regression problems. Specifically, one may draw inspiration from existing works such as the ones by~\citet{drucker1997improving, solomatine2004adaboost} who proposed boosting techniques for regression problems.

\revised{\paragraph{Adaptive Penalty Parameter.} A limitation of the adaptive choice of $\rho$ in~\cref{alg:penex_training} is that it does not retain the identification of the global additive logit offset established for the fixed-$\rho$ objective in~\cref{prop:self_calibration}. In the idealized setting where $\rho$ is recomputed exactly at every step without EMA, a common shift of all logits only changes the magnitude, but not the direction, of the resulting gradient. Consequently, the adaptive dynamics identify the logits only up to a common additive constant. This ambiguity does not affect the margin-tail and generalization results~(\cref{theorem:margin_max} and~\cref{cor:generalization_bound}), because they are invariant to a constant logit shift. An experiment with a tuned and fixed $\rho$ in~\cref{appx:fixed_rho} furthermore demonstrates that, if tuned, a fixed $\rho$ can lead to even better results. The adaptive algorithm~(\cref{alg:penex_training}) remains a competitive heuristic if tuning is not feasible.}

\section{Conclusion}\label{sec:conclusion}

We introduced the \textit{penalized exponential loss} (\ourmethod{}), a margin-promoting objective inspired by multi-class AdaBoost and designed for neural network training. We provided practical and theoretical motivation for its form and showed empirically that \ourmethod{} acts as a training regularizer: across various settings, it improves generalization over standard cross-entropy and the traditional formulation of the exponential loss. In low-data regimes, \ourmethod{} often surpasses established regularization techniques like label smoothing. {Beyond offering a practical loss function, our results also highlight a broader takeaway: in the form presented here, the exponential loss can be useful beyond the setting of the AdaBoost algorithm. In particular, \ourmethod{} demonstrates that the margin-emphasizing behavior that underpins AdaBoost can be translated into modern neural optimization. {Additionally, \ourmethod{} challenges the dominant paradigm in which a loss function is constructed as a well-behaved log-likelihood (e.g., cross-entropy) plus a separate “regularization term,” because its exponential form \textit{intrinsically regularizes} by encouraging large margins, rather than treating regularization as an add-on to a base log-likelihood.} Overall, we hope \ourmethod{} provides a simple and practical alternative to cross-entropy, particularly in limited-data regimes, and that it encourages further work on exponential-loss objectives beyond classical boosting.}

\begin{ack}
Michael Muehlebach is supported by the German Research Foundation (DFG) through the Emmy Noether Programme (project no. 456587626). The authors acknowledge support from the FWF Cluster of Excellence BilAI. Klaus-Rudolf Kladny acknowledges support from the International Max Planck Research School for Intelligent Systems (IMPRS-IS).
\end{ack}

\bibliography{main}
\bibliographystyle{tmlr}

\newpage

\appendix

\addcontentsline{toc}{section}{Appendix}
\vspace*{\fill}
{ \centering\part{{\huge{Appendix}}} \parttoc }
\vspace*{\fill}
\newpage

\section{Relationship to Multi-Class AdaBoost} \label{appx:rel_to_adaboost}

The essential ingredient of AdaBoost are so-called weak learners $g_m$, for all $m \in [M]$ that constitute a strong learner $f_M$ when considering their linear combination
{
\begin{equation*}
    f^{(y)}_M(\xb) \; = \; \sum_{m=1}^M \eta_m g^{(y)}_m(\xb),
\end{equation*}
for class $y \in [K]$}. These weak learners $g_m$, all together with their linear factors $\eta_m$, are fitted in a greedy fashion\footnote{Also known as \textit{stagewise additive modeling using a multi-class exponential loss function} (SAMME).} according to the CONEX objective~\cref{eq:L_CEL}
\begin{equation} \label{eq:adaboost_objective}
(g_m, \eta_m) \; \gets \; 
\arg \min_{g \in \Gcal_{\mathrm{\scriptscriptstyle CONEX}}, \; \eta > 0} \; \el \Big( f_{m-1} + \eta g; \, \alpha \Big),
\end{equation}
where $\alpha = (K-1)^{-1}$ and $\Gcal_{\scriptscriptstyle \mathrm{CONEX}}$ is the class of functions
\begin{align*}
    \Gcal_{\mathrm{\scriptscriptstyle CONEX}} \; \coloneqq \; \big\{ g : \Xcal \to \{\eb_1, \dots, \eb_K\} \big\}, \\
    \eb_i \;:=\; \big( \hat{\yb}^{(1)}, \dots, \hat{\yb}^{(K)} \big), \hat{\yb}^{(i)} = 1, \\
    \hat{\yb}^{(j)} = -(K-1)^{-1} \;\; (j \neq i),
\end{align*}
which ensures that the linear constraint $\sum_j f^{(j)}_m (\xb) = \sum_j \hat{\yb}^{(j)} = 0$ is fulfilled, for all $\xb \in \Xcal$~\cref{eq:L_CEL}. In practice, the $g_m$ are typically represented by decision stumps (e.g.,~\citet[p.362]{hastie2017elements}). For completeness, we show the full AdaBoost algorithm in~\cref{appx:SAMME}.

\paragraph{PENEX is Implicit AdaBoost.} We now proceed to demonstrate formally that gradient descent on \ourmethod{} is analogous to AdaBoost in that gradient descent steps { on a neural network parameterization $f_\theta$ with parameters $\theta$ } can be thought of as constrained parameters of weak learners that minimize the exponential loss. To see this, we consider a linearization about the previous parameters $\thetam{m-1}$, which assumes additivity as an approximation:
\begin{equation} \label{eq:first_order_approx}
    f_{\theta_{m}} (\xb) \; \approx \; f_{\theta_{m-1}} (\xb) \; + \; \eta \Jb(\xb) \Dthetam{m},
\end{equation}
where $\Jb(\xb)$ denotes the Jacobian of $f_{\theta_{m-1}}$ with respect to $\theta_{m-1}$ at $\xb$. This approximation~\cref{eq:first_order_approx} gives rise to an optimization problem closely related to that of AdaBoost~\cref{eq:adaboost_objective}, given as
\begin{equation} \label{eq:incremental_optimization_penex}
    \Dthetam{m}(\eta) \; \gets \; \arg \min_{\Dthetam{}:g_{\Dthetam{}} \in \Gcal_{\mathrm{PENEX}}} \el \left( f_{\theta_{m-1}} + \eta g_{\Dthetam{}}; \; \alpha \right),
\end{equation}
where $\Gcal_{\scriptscriptstyle \mathrm{PENEX}}$ is the set of functions given in~\cref{appx:gradient_weak_learner}. The following proposition shows that a small gradient descent step on \ourmethod{} indeed does solve~\cref{eq:incremental_optimization_penex}:
\begin{proposition} \label{prop:gradient_weak_learner}
In the limit $\eta \rightarrow 0$, for some $\Gcal_{\mathrm{\scriptscriptstyle PENEX}}$ and $\rho > 0$, the solution of~\cref{eq:incremental_optimization_penex} converges to the negative rescaled gradient of \ourmethod{}. That is,
\begin{equation*}
    \lim_{\eta \rightarrow 0} \Dthetam{m}(\eta) \; \propto \; -\nabla_\theta \pel (\thetam{m-1}; \, \alpha).
\end{equation*}
\end{proposition}
A proof for~\cref{prop:gradient_weak_learner} is provided in~\cref{appx:gradient_weak_learner}.

\section{Proofs} \label{appx:proofs}

\subsection[Proof for Proposition~\ref*{prop:self_calibration}]%
{Proof for \cref{prop:self_calibration}} \label{appx:fisher_consistency}

\begin{proof}
    The population-level equivalent of~\cref{eq:PEL}, conditionally on $\xb$, is
    \begin{equation} \label{eq:conditional_PEL_minimizer}
        \pel^{y|\xb}(f(\xb)) \; \coloneqq \; \EE_{y | \xb} \Big[ \exp \Big\{-\alpha f^{(y)}(\xb) \Big\} \Big] \; + \; \rho \sum_{j=1}^K \exp \Big\{ f^{(j)}(\xb) \Big\}.
    \end{equation}
    We note that~\cref{eq:conditional_PEL_minimizer} is strictly convex in $f(\xb)$ and differentiable. Hence, the unique minimum of~\cref{eq:conditional_PEL_minimizer}, called $\fopt(\xb)$, is characterized by its gradient being zero:
    \begin{equation}
        \frac{\partial \pel^{y|\xb}}{\partial \fopt^{(y)}(\xb)} \; = \; 0,
    \end{equation}
    which leads to
    \begin{equation*}
        P (y \, | \, \xb) \; = \; \frac{\rho}{\alpha} \exp \left\{ (1 + \alpha) \fopt^{(y)}(\xb) \right\}.
    \end{equation*}
    Due to re-normalization, the factor $\rho / \alpha$ is canceled out. Hence, for every $\xb$ that has non-zero support, the true conditional distribution $P (y \, | \, \xb)$ is recovered.
\end{proof}

\subsection[Proof for Proposition~\ref*{prop:gradient_weak_learner}]%
{Proof for \cref{prop:gradient_weak_learner}}\label{appx:gradient_weak_learner}
We choose the set of functions to be
\begin{equation} \label{eq:function_class_penex}
    \Gcal_{\scriptscriptstyle \mathrm{PENEX}} \; \coloneqq \; \Bigg\{ \Jb(\xb) \Dthetam{} \; \Big| \; \EEemp \Big[ \sum_{j=1}^K \exp \left\{ f_{\thetam{m-1}}^{(j)}(\xb) + \eta (\Jb(\xb) \Dthetam{m})^{(j)} \right\} \Big] \leq \epsilon, \; \norm{\Dthetam{}}_2 = 1 \Bigg\},
\end{equation}
where $\epsilon \, \coloneqq \, \EEemp [ \sum_{j=1}^K \exp \{ f_{\thetam{m-1}}^{(j)}(\xb) \} ]$.

\begin{proof}
    We start by performing a Taylor approximation of the objective function and the constraints around $\eta=0$ in~\cref{eq:incremental_optimization_penex}. This yields
    \begin{equation*}
        \EEemp \left[ \exp\left\{ - \alpha f_{\theta_{m-1}}^{(y)}(\xb)\right\}\right] -\alpha \eta \EEemp \left[ \exp\left\{ - \alpha f_{\theta_{m-1}}^{(y)}(\xb)\right\} \Jb(\xb)^{(y)}\right] \Dthetam{}+h_1(\eta,\Dthetam{}) \eta
    \end{equation*}
    for the objective function and
\begin{equation*}
\EEemp \Big[ \sum_{j=1}^K \exp \left\{ f_{\thetam{m-1}}^{(j)}(\xb) \right\} \Big] + \eta \EEemp \Big[ \sum_{j=1}^K \exp \left\{ f_{\thetam{m-1}}^{(j)}(\xb) \right\} \Jb(\xb)^{(j)} \Big] \Dthetam{} + h_2(\eta,\Dthetam{}) \eta \leq \varepsilon
\end{equation*}
for the constraints, where $h_1(\eta,\Dthetam{})$, $h_2(\eta,\Dthetam{})$ are continuous functions that satisfy $\lim_{\eta\rightarrow 0} h_1(\eta,\Dthetam{})=0$ and $\lim_{\eta\rightarrow 0} h_2(\eta,\Dthetam{})=0$, and where, by assumption, 
\begin{equation*}
    \epsilon=\EEemp \Big[ \sum_{j=1}^K \exp \{ f_{\thetam{m-1}}^{(j)}(\xb) \} \Big].
\end{equation*} 
In addition, we used the notation $\Jb(\xb)^{(j)}$ to refer to the $j$th row of the Jacobian matrix $\Jb(\xb)$. The optimization in~\cref{eq:incremental_optimization_penex} can therefore be rewritten as
\begin{align}
v(\eta) \quad \coloneqq \quad \min_{\Dthetam{}} -\alpha &\EEemp \left[ \exp\left\{ - \alpha f_{\theta_{m-1}}^{(y)}(\xb)\right\} \Jb(\xb)^{(y)}\right] \Dthetam{}+h_1(\eta,\Dthetam{}) \label{eq:prooftmp}\\
\text{subject to}\quad &\EEemp \Big[ \sum_{j=1}^K \exp \left\{ f_{\thetam{m-1}}^{(j)}(\xb) \right\} \Jb(\xb)^{(j)} \Big] \Dthetam{} + h_2(\eta,\Dthetam{}) \leq 0 \nonumber\\
& \norm{\Dthetam{}}_2^2 = 1. \nonumber
\end{align}
For small $\eta$, the first constraint reduces to a half-space passing through the origin. This implies that the constraint can be strictly satisfied for $\eta=0$ and therefore, due to continuity, the feasible set in~\cref{eq:incremental_optimization_penex} (or equivalently~\cref{eq:prooftmp}) is nonempty for small enough $\eta$. We further note that $v(\eta)$, the minimum value in~\cref{eq:prooftmp}, is continuous for small $\eta$, see for example~\citep[][Ch.~5]{stillparametric2018} and consider any sequence $\eta_k>0$ and corresponding $\Dthetam{mk}$, $k = 1, 2, \dots$ where $\eta_k\rightarrow 0$ and $\Dthetam{mk}$ belongs to the corresponding set of minimizers of~\cref{eq:incremental_optimization_penex}. Our goal is to show that any limit point of the sequence belongs to the set of minimizer of~\cref{eq:first_order_approx} with $\eta = 0$. To that extend, we first note that $\Dthetam{mk}$ remains bounded, since $\norm{\Dthetam{mk}}_2^2=1$. Consider any limit point $\Dthetam{\infty}$ of $\Dthetam{mk}$ and pass to a subsequence $\Dthetam{mk}$ such that $\Dthetam{mk}\rightarrow\Dthetam{\infty}$. We now show that $\Dthetam{\infty}$ is a minimizer of~\cref{eq:incremental_optimization_penex} (or equivalently~\cref{eq:prooftmp}) for $\eta=0$. The limit point $\Dthetam{\infty}$ satisfies $\norm{\Dthetam{\infty}}_2^2=1$ and the first constraint
\begin{equation*}
    \EEemp \Big[ \sum_{j=1}^K \exp \left\{ f_{\thetam{m-1}}^{(j)}(\xb) \right\} \Jb(\xb)^{(j)} \Big] \Dthetam{\infty} \leq 0,
\end{equation*}
and is therefore feasible for the minimization in~\cref{eq:prooftmp} when $\eta=0$. We conclude from the continuity of $v$ that $v(\eta_k) \rightarrow v(0)$ and therefore $\Dthetam{\infty}$ is a minimizer of \eqref{eq:incremental_optimization_penex} for $\eta=0$. Hence the entire sequence $\Dthetam{mk}$ converges to the set of minimizers of \eqref{eq:incremental_optimization_penex} for $\eta=0$. As a result, $\Dthetam{\infty}$ satisfies the Karush-Kuhn-Tucker conditions corresponding to \eqref{eq:prooftmp} for $\eta=0$. These can be expressed as
\begin{equation*}
\Dthetam{\infty} \propto \alpha \EEemp \Big[ \sum_{j=1}^K \exp \left\{ -\alpha f_{\thetam{m-1}}^{(j)}(\xb) \right\} \Jb(\xb)^{(j)} \Big] - \bar{\rho} \EEemp \Big[ \sum_{j=1}^K \exp \left\{ f_{\thetam{m-1}}^{(j)}(\xb) \right\} \Jb(\xb)^{(j)} \Big],
\end{equation*}
where $\bar{\rho}\geq 0$ is related to a multiplier of the first constraint in~\cref{eq:prooftmp}. This means that $\Dthetam{\infty} \propto - \nabla_\theta \pel (\theta_{m-1}; \alpha)$ with $\rho = \bar{\rho}$, which concludes the proof.   

\subsection[Proof for Theorem~\ref*{theorem:margin_max}]%
{Proof for \cref{theorem:margin_max}} \label{appx:margin_max}

We first apply a Chernoff bound to see that
\begin{equation*}
     \PP (m_{f}(\xb, y) \leq \gamma) \, \leq \, \EE \left[ \exp \left\{ \beta \left( \gamma + \max_{j \neq y} f^{(j)}(\xb) - f^{(y)}(\xb) \right) \right\} \right],
\end{equation*}
for any $\beta \in (0, 1)$. Then, we note that for any $(\xb, y)$
\begin{equation*}
     \max_{j \neq y} f^{(j)}(\xb) \; \leq \; \log \left( \sum_{j=1}^K \exp \left\{ f^{(j)}(\xb) \right\} \right),
\end{equation*}
which we combine with Hölder's inequality for $\alpha \coloneqq \frac{\beta}{1 - \beta}$\footnote{We recall that $\beta \in (0, 1)$, so $\frac{1}{\beta} \geq 1$ and $\frac{1}{1 - \beta} \geq 1$, which is a condition for Hölder's inequality.} to see that
\begin{equation*}
     \Prob (m_{f}(\xb, y) \leq \gamma) \; \leq \; e^{\beta \gamma} \EE[\el(f; \, \alpha)]^{1 - \beta} \cdot \EE[\mathrm{SE}(f)]^\beta.
\end{equation*}
We can now apply the weighted AM-GM inequality to get
\begin{equation}
    \EE [\el(f; \, \alpha)]^{1 - \beta} \cdot (\rho \EE[\mathrm{SE}(f)])^\beta \rho^{-\beta}\; \leq \; ((1 - \beta) \EE[\el(f; \, \alpha)] + \beta \rho \EE [ \mathrm{SE}(f) ]) \rho^{-\beta}
\end{equation}
and further, because $\beta \in (0, 1)$,
\begin{equation*}
    (1 - \beta) \EE [ \el(f; \, \alpha) ] + \beta \rho \EE [ \mathrm{SE}(f) ] \; \leq \; \EE [\el (f; \, \alpha)] + \rho \EE [ \mathrm{SE}(f) ] \; = \; \EE [\pel(f; \, \alpha, \rho)].
\end{equation*}
Thus, we have
\begin{equation*}
     \Prob (m_{f}(\xb, y) \leq \gamma) \; \leq \; e^{\beta \gamma} \rho^{- \beta} \EE [\pel(f; \, \alpha, \rho)].
\end{equation*}
The result follows from replacing $\beta$ by $\beta = \frac{\alpha}{\alpha + 1}$.
\end{proof}

\section[Differences to Schapire et al. (1998)]%
{{Differences Between~\cref{theorem:margin_max} and the Result by~\citet{bartlett1998boosting}}} \label{appx:margin_result_difference}

The earlier work by~\citet{bartlett1998boosting} shows that AdaBoost effectively increases margins. We recapitulate the Theorem 5 of~\citet{bartlett1998boosting}, using the notation from the present work:
\begin{theorem}[Theorem 5 of~\citet{bartlett1998boosting}] \label{theorem:earlier_margin_maximiation_result}
    Consider a binary classification problem ($K = 2$) with $f^{(2)}(\xb) \equiv -f^{(1)}(\xb)$ and assume that $f^{(1)}(\xb) \, = \, \sum_{m=1}^M \eta_m g_m(\xb)$ with weighted training errors
    \begin{equation*}
        \epsilon_m \, \coloneqq \, \PP_{(\xb_i, y_i) \sim \Phat_m} (\indfonearg{f^{(y_i)}(\xb_i) < 0}),
    \end{equation*}
    where the probability mass of the weighted empirical distribution $\Phat_m$ is given as
    \begin{equation}
        \Phat_m(\xb_i, y_i) \, \coloneqq \, \frac{\Phat_{m-1}(\xb_i, y_i) \exp \{ - \eta_m f_m^{(y_i)} (\xb_i) \}}{Z_m} 
    \end{equation}
    and
    \begin{equation}
        \eta_m \coloneqq \frac{1}{2} \log \left( \frac{1 - \epsilon_m}{\epsilon_m} \right)
    \end{equation}
    with uniform initialization $\Phat_1 = \Ucal([K])$. Then, for any $\gamma > 0$, it holds that 
    \begin{equation*}
        \PPast(f_M^{(y)}(\xb) \, \leq \, \gamma) \, \leq \, 2^{M} \prod_{m=1}^M \sqrt{\epsilon_m^{1 - \gamma} (1 - \epsilon_m)^{1 + \gamma}},
    \end{equation*}
    for $f_M$ fitted via binary AdaBoost~\citep{freund1995desicion}.
\end{theorem}
A generalization of~\cref{theorem:earlier_margin_maximiation_result} to multi-class AdaBoost has been provided by~\citet{wang2013boosting}. These earlier results differ from our result~(\cref{theorem:margin_max}) in three fundamental ways\footnote{We only refer to~\cref{theorem:earlier_margin_maximiation_result} from hereon, because the same arguments can be made for the results presented by~\citet{wang2013boosting}}:

\begin{enumerate}
    \item \cref{theorem:earlier_margin_maximiation_result} makes a strong structural assumption about the classifier and provides a bound in terms of the weighted training errors of the individual ``intermediate learners'' $f_m$. Our result, in contrast, makes no assumptions about $f$.
    \item \cref{theorem:earlier_margin_maximiation_result} assumes that $f$ is fitted via the AdaBoost algorithm.~\cref{theorem:margin_max} makes no assumption about how $f$ is trained.
    \item We consider a new formulation of the exponential loss that differs from the standard exponential loss in that the class scores/logits $f^{(y)}$ do not need to sum up to zero (concretely, the $f^{(2)}(\xb) \equiv -f^{(1)}(\xb)$ assumption in~\cref{theorem:earlier_margin_maximiation_result}). Instead, PENEX incorporates a penalty to prevent the logits from diverging (see~\cref{eq:PEL}). The margin result from~\cref{theorem:earlier_margin_maximiation_result} cannot be directly translated to the PENEX formulation.
\end{enumerate}

\section{Stagewise Additive Modeling using a Multi-class Exponential loss} \label{appx:SAMME}

The standard multi-class AdaBoost solves~\cref{eq:adaboost_objective} via a specific algorithm called SAMME, shown in~\cref{alg:SAMME}.

\begin{algorithm}[H]
 \SetAlgoLined
 \KwIn{Training set $\{ (\xb_i, y_i) \}_{i=1}^n$, where $\xb_i \in \mathcal{X}$ and $y_i \in [K]$}
 \KwOut{Strong classifier $f_M$}
 Initialize sample weights $w_1(i) \gets \frac{1}{n}$ for $i=1, \ldots, n$ \\
 \For{$m = 1$ \KwTo $M$}{
    Fit weak classifier $g_m$ using weights $w_m$ \\
    Compute weighted error $\epsilon_m \gets \sum_{i=1}^n w_m(i) \cdot \indfonearg{g_m(\xb_i) \neq y_i}$ \\
    Compute classifier weight $\eta_m \gets \log \left( \frac{1 - \epsilon_m}{\epsilon_m} \right) \, + \, \log(K - 1)$ \\
    Update sample weights:
    \[
    w_{m+1}(i) \; \gets \; \frac{w_m(i) \exp(-\eta_m \indfonearg{ g_m(\xb_i) \neq y_i} )}{Z_m},
    \]
    where $Z_m$ is the partition function
 }
 Combine weak classifiers: $f_M(\xb) = \arg\max_{k \in [K]} \sum_{m=1}^M \eta_m g_m(\xb)$
 \caption{SAMME (Multi-class AdaBoost)} \label{alg:SAMME}
\end{algorithm}

\section{Practical Implementation of \ourmethod{}} \label{appx:penex_practical}

\subsection{Inference} \label{appx:penex_inference}

Motivated by~\cref{prop:self_calibration}, we rescale logits that are optimized on PENEX by $1 + \alpha$. Specifically, to obtain a prediction $\Phat_{\scriptscriptstyle \textrm{PENEX}}$, we always use
\begin{equation}
    \Phat_{\scriptscriptstyle \textrm{PENEX}}(y \, | \, \xb) \; \coloneqq \; \frac{\exp \{ (1 + \alpha) f^{(y)}(\xb) \}}{\sum_{j=1}^K \exp \{ (1 + \alpha) f^{(j)}(\xb) \}},
\end{equation}
where $f^{(j)}(\xb)$ are the logits obtained by optimizing $\pel$~\cref{eq:PEL}. 

\subsection{Default Value for Sensitivity $\alpha$}

While the optimal choice of sensitivity $\alpha$ depends on various factors such as training set size, model architecture, amount of classes and label noise, we notice that the tuned value for $\alpha$ is typically close to
\begin{equation} \label{eq:default_alpha}
    \alpha \; = \; 0.1.
\end{equation}
If~\ourmethod{} is used without tuning, we generally recommend~\cref{eq:default_alpha} as the default choice. If, however, there is substantial label noise, we recommend a smaller value. If there is little label noise, $\alpha$ may be chosen larger, leading to more aggressive margin increase.

\subsection{Default Values for~\cref{alg:penex_training}}

For all experiments conducted in the present work, we use the following hyper parameter configuration:

\begin{table}[h]
\begin{center}
\begin{adjustbox}{max width=\textwidth}
\begin{tabular}{@{}lc@{}}
\toprule
\textbf{Parameter} & \textbf{Value} \\ \midrule
EMA factor $\beta$ & $0.1$ \\ \midrule
Lower penalty bound $\rho_{\mathrm{min}}$  & $10^{-6}$ \\ \midrule
Upper penalty bound $\rho_{\mathrm{max}}$ & $100.0$ \\ \midrule
\end{tabular}
\end{adjustbox}
\end{center}
\end{table}

In practice, we observe in all conducted experiments that results typically do not differ much if these parameters are chosen differently.

\subsection{PyTorch-like Code}

As mentioned in~\cref{sec:practical_implementation}, \ourmethod{} does not require a custom training loop in order to implement~\cref{alg:penex_training}. A PyTorch-like loss module is displayed in~\cref{lst:pseudocode}, with only 34 lines of code (including comments). It is easy to see that it can be integrated seamlessly into any standard training loop.

\begin{listing} \label{lst:pseudocode}
\begin{codebox}
\begin{minted}{python}
import torch
import torch.nn as nn


def exp_loss(logits, targets, alpha):
    """Exponential loss on the true-class score fy(x)."""
    y  = targets.long()
    fy = logits.gather(1, y.unsqueeze(1)).squeeze(1)
    return torch.exp(-alpha * fy)

class PENEX(nn.Module):
    """Ultra-minimal readable version."""
    def __init__(self, alpha=0.1, rho_min=1e-6, rho_max=100.0, ema=0.1):
        super().__init__()
        self.alpha   = alpha
        self.rho_min = rho_min
        self.rho_max = rho_max
        self.ema     = ema
        self.rho     = None  # running estimate of rho

    def forward(self, logits, targets):
        # Base loss
        base   = exp_loss(logits, targets, self.alpha)
        # Penalty term; uses torch.logsumexp() for numerical stability
        sumexp = torch.exp(torch.logsumexp(logits, dim=1))

        # Update rho without tracking gradients
        with torch.no_grad():
            est = self.alpha * (base.mean() / (sumexp.mean() + 1e-12))
            est = est.clamp(self.rho_min, self.rho_max)
            self.rho = est if self.rho is None else (1 - self.ema) * self.rho + self.ema * est

        loss = base + self.rho * sumexp
        return loss.mean()
\end{minted}
\caption{Simplified PyTorch-like code for the PENEX loss.}
\end{codebox}
\end{listing}

\section{Experimental Details} \label{appx:experimental_details}

\subsection{Libraries}

For our implementation, we use \texttt{PyTorch Lightning}~\citep{Falcon_PyTorch_Lightning_2019}, a framework built on top of \texttt{PyTorch}~\citep{NEURIPS2019_9015}. For all experiments involving language models, we use the \texttt{transformers}, \texttt{datasets} and \texttt{evaluate} Hugging Face libraries. For hyperparameter tuning, we use the \texttt{Optuna} library~\citep{optuna_2019}. We use \texttt{SciencePlots}~\citep{SciencePlots} for appealing plots and \href{https://github.com/wandb/wandb}{\texttt{weights \& biases}} for development and analysis.

\subsection{Dataset \& Model Licenses} \label{appx:licenses}

The licenses of each dataset and pretrained model used in the present work are listed in the following table:

\begin{table}[h]
\begin{center}
\begin{adjustbox}{max width=\textwidth}
\begin{tabular}{@{}lc@{}}
\toprule
\textbf{Dataset} & \textbf{License} \\ \midrule
CIFAR-10 & MIT \\ \midrule
PathMNIST  & CC BY 4.0 \\ \midrule
BBC News & Apache 2.0 \\ \midrule
RoBERTa & MIT \\ \midrule
ImageNet & Non-commercial research/educational \\ \midrule
\end{tabular}
\end{adjustbox}
\end{center}
\end{table}

\newpage

\subsection{Computational Resources} \label{appx:computational_ressources}

All experiments are run on a \href{https://htcondor.org/htcondor/overview/}{HTCondor cluster}. All models are trained with a single GPU and CPU (except for ImageNet, which is trained with $4$ GPUs and $32$ CPUs), where we adjust the type of GPU for each experiment. Specifically, all experiments can be run on the following hardware:

\begin{table}[h]
\begin{center}
\begin{adjustbox}{max width=\textwidth}
\begin{tabular}{@{}lc@{}}
\toprule
\textbf{Dataset} & \textbf{GPU} \\ \midrule
CIFAR-10 & NVIDIA A100-SXM4-40GB \\ \midrule
Noisy CIFAR-10 & NVIDIA A100-SXM4-40GB \\ \midrule
CIFAR-100 & NVIDIA A100-SXM4-40GB \\ \midrule
PathMNIST & NVIDIA A100-SXM4-40GB \\ \midrule
BBC News & NVIDIA A100-SXM4-40GB \\ \midrule
ImageNet & NVIDIA A100-SXM4-80GB \\ \midrule
\end{tabular}
\end{adjustbox}
\end{center}
\end{table}
We note, however, that all listed results should be achievable on different hardware (as we do not measure properties like runtime). In addition, all experiments are run with a single worker (as recommended by PyTorch for the used setup). 

\newpage

The amount of CPU memory also differs between experiments:
\begin{table}[h]
\begin{center}
\begin{adjustbox}{max width=\textwidth}
\begin{tabular}{@{}lc@{}}
\toprule
\textbf{Dataset} & \textbf{Memory (in megabytes)} \\ \midrule
CIFAR-10 & 8096 \\ \midrule
Noisy CIFAR-10 & 8096 \\ \midrule
CIFAR-100 & 16192 \\ \midrule
PathMNIST & 8096 \\ \midrule
BBC News & 16192 \\ \midrule
ImageNet & 150000 \\ \midrule
\end{tabular}
\end{adjustbox}
\end{center}
\end{table}

We furthermore report the approximate runtime for a single run of an experiment (200 epochs):
\begin{table}[h]
\begin{center}
\begin{adjustbox}{max width=\textwidth}
\begin{tabular}{@{}lc@{}}
\toprule
\textbf{Dataset} & \textbf{Approx. time per run (in minutes)} \\ \midrule
CIFAR-10 Training & 60 \\ \midrule
CIFAR-100 Training & 120 \\ \midrule
PathMNIST Training & 18 \\ \midrule
BBC News Training & 60 \\ \midrule
ImageNet Training & 3600 \\ \midrule
\end{tabular}
\end{adjustbox}
\end{center}
\end{table}

\subsection{Ablation Implementations} \label{appx:baselines}

To take the zero-sum constraint in CONEX~\cref{eq:L_CEL} into account, we first define the constraint
\begin{equation*}
    h(\xb; \, \theta) \coloneqq \oneb^{\top} f_\theta (\xb)
\end{equation*}
and see that
\begin{equation*}
    h(\xb; \, \theta) = 0, \; \forall \xb \in \text{supp}(P) \; \iff \; \EE [ h(\xb; \, \theta)^2 ] = 0.
\end{equation*}
\paragraph{Squared Penalty.} We penalize deviations from the penalty constraint via a quadratic penalty. To this end, we construct the loss function
\begin{equation*}
    \Lcal (\thetam{m}) \; = \; \el(\thetam{m}) + \frac{\rho}{2} \Big( \EEemp \left[ h(\xb; \, \thetam{m})^2 \right] \Big)^2,
\end{equation*}
the optimum of which converges to the constrained solution for $\rho \rightarrow \infty$ (for a fixed learning rate $\eta$). In practice, however, choosing large $\rho$ leads to an ill-conditioned loss surface, which is why a trade-off must be made in the choice of $\rho$.

\paragraph{Augmented Lagrangian.} The augmented Lagrangian method~\citep{hestenes1969multiplier} considers an objective that arises from~\cref{eq:L_CEL} by introducing a penalty term with parameter $\rho > 0$ and the dual variable $\lambda_m \in \RR$:
\begin{equation*}
    \Lagr (\thetam{m}) \; \coloneqq \; \el(\thetam{m}) \, + \, \frac{\rho}{2} \Big( \EEemp \left[ h(\xb; \, \thetam{m})^2 \right] \Big)^2 \, + \, \lambda_m \Big( \EEemp \left[ h(\xb ; \, \thetam{m})^2 \right] \Big).
\end{equation*}
The algorithm updates the primal ($\thetam{m}$) and dual ($\lambda_m$) variables in an alternating fashion:
\begin{equation*}
    \thetam{m} \gets \thetam{m-1} - \eta \nabla_{\theta} \Lagr (\thetam{m-1}), \qquad \lambda_m \gets \lambda_{m-1} \; + \; \frac{\rho}{\nu} \Big( \EEemp \left[ h(\xb; \, \thetam{m-1})^2 \right] \Big),
\end{equation*}
where $\nu > 0$ is an inverse scaling factor.

\underline{Note}: In practice, computing $\EEemp$ would require loading the entire training set into memory. We thus resort to sub-sampling $\EEemp$ from batches. 

\subsection{Metrics} \label{appx:metrics}

\paragraph{Accuracy.} We measure the accuracy of a classifier $f$ as
\begin{equation*}
    \text{ACC}(f) \; \coloneqq \; \frac{1}{N} \sum_{i=1}^N \mathbbm{1} \Big\{ y_i \; = \; \arg \max_j f^{(j)}(\xb_i) \Big\}.
\end{equation*}

\paragraph{Expected Calibration Error.} Formally, it is given as
\begin{equation*}
    \text{ECE}(f) \; \coloneqq \; \sum_{m=1}^M \frac{|B_m|}{n} \big| \text{ACC}(f; \, B_m) - \text{CONF} (f; \, B_m) \big|,
\end{equation*}
where $B_m$ is the set of data points whose prediction confidence falls into the interval 
\begin{equation*}
    I_m \; \coloneqq \; \Big(\frac{m-1}{M}; \, \frac{m}{M} \Big]
\end{equation*}
and
\begin{equation*}
    \text{CONF}(f) \; \coloneqq \; \frac{1}{N} \sum_{i = 1}^N \Phat (y_i \; | \; f(\xb_i)).
\end{equation*}
We use the default $m=15$, as implemented by the \href{https://lightning.ai/docs/torchmetrics/stable/classification/calibration_error.html}{torchmetrics} library.

\paragraph{Brier Score.} We compute the Brier score as
\begin{equation*}
    \text{BRIER}(f) \; \coloneqq \; \frac{1}{N} \sum_{i=1}^N \sum_{j=1}^K \left( \Phat (y = j \; | \; f(\xb_i)) - \yb^{(j)}_i \right)^2,
\end{equation*}
where
\begin{equation*}
    \yb^{(j)}_i \; \coloneqq \;  \begin{cases} 1, \quad \text{if} \; \; j = y_i \\ 0, \quad \text{else,} \end{cases}
\end{equation*}
is a one-hot encoding of the label.

\subsection{Data Preprocessing}

\subsubsection{Computer Vision}

We do not perform data augmentation for any computer vision task, except for ImageNet.

\paragraph{CIFAR-10/CIFAR-100/PathMNIST.} We perform standard normalization for these datasets, with the following mean and standard deviation values
\begin{equation*}
\mu = (0.5, 0.5, 0.5), \; \sigma = (0.5, 0.5, 0.5),   
\end{equation*}
where we have three values for each parameter because standardization is applied per channel (there are three channels). For PathMNIST in particular, prior to standardization, we load the dataset at a resolution of $64 \times 64$ and further scale it down to $32 \times 32$.

\paragraph{ImageNet.} We also perform normalization, but instead with the following (standard) values
\begin{equation*}
\mu = (0.485,0.456,0.406), \; \sigma = (0.229,0.224,0.225).   
\end{equation*}
In addition, as common with vision transformers, we apply the following data augmentations: images are first resized to 256 pixels along the shorter side, then randomly cropped to $224 \times 224$ pixels with a scale factor uniformly sampled between $0.8$ and $1.0$, followed by a random horizontal flip with probability $0.5$.

\clearpage

\subsubsection{Fine-Tuning RoBERTa}

There exist five labels in total corresponding to the categories tech, business, sport, entertainment and politics. To prevent memory overload, the input length is limited to a maximum length of $512$ tokens and text is clipped if this length is exceeded (this is fairly common in other works, e.g.,~\citet{kladny2025conformal}). An example text input for the label \textit{business} is demonstrated in the following:

\begin{longtable}{p{0.9\textwidth}}
    \toprule
    \textbf{Input} \\
    \midrule
    \endfirsthead

    \toprule
    \textbf{Prompt (continued)} \\
    \midrule
    \endhead
\begin{lstlisting}[
        breakatwhitespace=true,  %
        breaklines=true
    ]
china aviation seeks rescue deal scandal-hit jet fuel supplier china aviation oil has offered to repay its creditors $220m (117m) of the $550m it lost on trading in oil futures.  the firm said it hoped to pay $100m now and another $120m over eight years. with assets of $200m and liabilities totalling $648m  it needs creditors  backing for the offer to avoid going into bankruptcy. the trading scandal is the biggest to hit singapore since the $1.2bn collapse of barings bank in 1995. chen jiulin  chief executive of china aviation oil (cao)  was arrested by at changi airport by singapore police on 8 december. he was returning from china  where he had headed when cao announced its trading debacle in late-november. the firm had been betting heavily on a fall in the price of oil during october  but prices rose sharply instead.  among the creditors whose backing cao needs for its restructuring plan are banking giants such as barclay s capital and sumitomo mitsui  as well as south korean firm sk energy. of the immediate payment  the firm - china s biggest jet fuel supplier - said it would be paying $30m out of its own resources. the rest would come from its parent company  china aviation oil holding company in beijing. the holding company  owned by the chinese government  holds most of cao s singapore-listed shares. it cut its holding from 75% to 60% on 20 october.
\end{lstlisting} \\
    \bottomrule
\end{longtable}

In addition to clipping, we sub-sample a training set of size $200$ to enhance the difficulty of the classification task.

\subsection{Train/Validation Splits}

In our experiments, we always construct an additional validation set from the training set. We do this in order to ensure that the test task provides an unbiased estimate of the desired model scores. The split ratios are determined by the size of the training set (we choose a larger ratio for ImageNet, because this data set is very large) provided in the following table:

\begin{table}[h]
\begin{center}
\begin{adjustbox}{max width=\textwidth}
\begin{tabular}{@{}lcccccccc@{}}
\toprule
\textbf{Method} & Training/Validation Split Ratio \\ \midrule
CIFAR-10  &  0.8 \\ \midrule
CIFAR-100 & 0.8 \\ \midrule
PathMNIST  & 0.8 \\ \midrule
ImageNet  & 0.9 \\ \midrule
BBC News  & 0.8 \\
\bottomrule
\end{tabular}
\end{adjustbox}
\end{center}
\end{table}

\clearpage

\subsection{Architectures} \label{appx:architectures}

We generally do not apply much regularization within the architectures (i.e., we do not use batch normalization) in order to better compare the effects of the used loss-based regularizers.

\paragraph{CIFAR-10/100 \& PathMNIST (experiments from the main text).} The architecture used is provided in the following table:

\begin{center}
\begin{tabular}{lll}
\toprule
\textbf{Layer Type} & \textbf{Parameters} & \textbf{Output Shape} \\
\midrule
Input & - & $3 \times 32 \times 32$ \\
Conv2d & 32 filters, $3\times3$, padding = $1$ & $32 \times 32 \times 32$ \\
ReLU & - & $32 \times 32 \times 32$ \\
MaxPool2d & $2 \times 2$, stride = $2$ & $32 \times 16 \times 16$ \\
Conv2d & $64$ filters, $3\times3$, padding = $1$ & $64 \times 16 \times 16$ \\
ReLU & - & $64 \times 16 \times 16$ \\
MaxPool2d & $2\times2$, stride = $2$ & $64 \times 8 \times 8$ \\
Conv2d & $128$ filters, $3\times3$, padding = $1$ & $128 \times 8 \times 8$ \\
ReLU & - & $128 \times 8 \times 8$ \\
MaxPool2d & $2\times2$, stride = $2$ & $128 \times c \times c$ \\
Flatten & - & $2048$ \\
Dropout & p = $0.5$ & $2048$ \\
Linear & $2048 \rightarrow 256$ & $256$ \\
ReLU & - & $256$ \\
Linear & $256 \rightarrow \texttt{n\_out}$ & \texttt{n\_out} \\
\bottomrule
\end{tabular}
\end{center}
In the final layer, $\texttt{n\_out}=10$, $\texttt{n\_out}=100$ or $\texttt{n\_out}=9$, respectively. We do so for all methods except for ablation 3)~(see~\cref{sec:experiments_setup}), where $\texttt{n\_out}=9$, $\texttt{n\_out}=99$ or $\texttt{n\_out}=8$, respectively. Furthermore, we note that $c=4$ for CIFAR-10/100 (because the images have shape $3 \times 32 \times 32$) and $c=8$ for PathMNIST (images have shape $3 \times 64 \times 64$).

\paragraph{{CIFAR-10 \& CIFAR-100 (additional ResNet experiment} in~\cref{appx:alternative_architecture_experiment}).}

\begin{center}
\begin{tabular}{lll}
\toprule
\textbf{Layer Type} & \textbf{Parameters} & \textbf{Output Shape} \\
\midrule
Input & -- & $3\times32\times32$ \\
Conv2d + BN + ReLU & $64$, $3\times3$, s=1, p=1 & $64\times32\times32$ \\
\midrule
\textbf{Layer 1: BasicBlock$\times2$} & $3\times3$ convs, stride $1$ & $64\times32\times32$ \\
\midrule
\textbf{Layer 2: BasicBlock$\times2$} & first block: s=2; shortcut $1\times1$ & $128\times16\times16$ \\
      & second block: stride $1$ & $128\times16\times16$ \\
\midrule
\textbf{Layer 3: BasicBlock$\times2$} & first block: s=2; shortcut $1\times1$ & $256\times8\times8$ \\
      & second block: stride $1$ & $256\times8\times8$ \\
\midrule
\textbf{Layer 4: BasicBlock$\times2$} & first block: s=2; shortcut $1\times1$ & $512\times4\times4$ \\
      & second block: stride $1$ & $512\times4\times4$ \\
\midrule
AvgPool2d & $4\times4$ global pool & $512\times1\times1$ \\
Flatten & -- & $512$ \\
Linear & $512\rightarrow 10$ & 10 \\
\bottomrule
\end{tabular}
\end{center}

\newpage

\paragraph{ImageNet.} The architecture used corresponds to ViT-Base~\citep{Dosovitskiy2021image} and is provided in the following table:

\begin{center}
\begin{tabular}{lll}
\toprule
\textbf{Layer Type} & \textbf{Parameters} & \textbf{Output Shape} \\
\midrule
Input & - & $3 \times 224 \times 224$ \\
Patch Embedding & Patch size $16\times16$, embedding dim $768$ & $196 \times 768$ \\
Add CLS token & - & $197 \times 768$ \\
Add Positional Encoding & - & $197 \times 768$ \\
Dropout & p = 0.1 & $197 \times 768$ \\
\midrule
\multicolumn{3}{c}{\textbf{Transformer Encoder (12 blocks)}} \\
\midrule
Multi-Head Self-Attention & heads = 12, dim\_head = 64 & $197 \times 768$ \\
Add \& Layer Norm & - & $197 \times 768$ \\
MLP & hidden dim = 3072, dropout = 0.1 & $197 \times 768$ \\
Add \& Layer Norm & - & $197 \times 768$ \\
(repeated 12 times) & & \\
\midrule
Pooling & CLS token (pool = 'cls') & $768$ \\
MLP Head & Linear $768 \rightarrow \texttt{n\_out}$ & \texttt{n\_out} \\
\bottomrule
\end{tabular}
\end{center}
In the final layer, $\texttt{n\_out} = 1000$.

\subsection{Training} \label{appx:training}

For all methods, we use either the Adam optimizer~\citep{kingma2014adam} or the AdamW optimizer~\citep{loshchilov2017decoupled} with default parameters. This means that both optimizer are used with $\beta_1 = 0.9$ and $\beta_2 = 0.999$. All parameters are given in the following table:
\begin{table}[h]
\begin{center}
\begin{adjustbox}{max width=\textwidth}
\begin{tabular}{@{}lcccccccc@{}}
\toprule
\textbf{Method} & Learning Rate & Optimizer & \shortstack{\# Training \\ Epochs} & \shortstack{Batch Size \\ (effective)}& \shortstack{Learning Rate \\ Scheduling} & \shortstack{Gradient Clip \\ Value} & \shortstack{\# Gradient \\ Acc. Steps} & Precision \\ \midrule
CIFAR-10  &        $1 \cdot 10^{-4}$         & Adam                  & 200                & 64 & constant & 5.0 & 1 & medium \\ \midrule
CIFAR-10~(\cref{appx:alternative_architecture_experiment}) &        $1 \cdot 10^{-1}$         & SGD                  & 200                & 128 & cosine & 5.0 & 1 & medium \\ \midrule
CIFAR-100 & $1 \cdot 10^{-4}$         & Adam                  & 200                & 64 & constant & 15.0 & 1 & medium \\ \midrule
PathMNIST  & $1 \cdot 10^{-4}$         & Adam                  & 200                & 64 & constant & 10.0 & 1 & medium \\ \midrule
ImageNet  & \shortstack{$1 \cdot 10^{-5}$ for \ourmethod{} \\ and $1 \cdot 10^{-3}$ for others}        & Adam                  & 300                & 2048 & \shortstack{cosine \& \\ $10$ epochs warmup} & 1.0 & 1 & medium \\ \midrule
BBC News  & $5 \cdot 10^{-6}$         & AdamW                  & 200                & 8 &\shortstack{linear \\ (no warmup)} & 5.0 & 2 & high \\ \midrule
\end{tabular}
\end{adjustbox}
\end{center}
\end{table}

\newpage

\subsection{Hyperparameter Tuning} \label{appx:hparams}

For each method, we use the tree-structured Parzen estimator with $50$ trials per method, using the standard parameters from the \texttt{Optuna} library. We list the parameter ranges for hyperparameter tuning that were used for each baseline:

\begin{table}[h]
\begin{center}
\begin{adjustbox}{max width=\textwidth}
\begin{tabular}{@{}llcc@{}}
\toprule
\textbf{Method} & \textbf{Hyperparameter} & \textbf{Lower Bound} & \textbf{Upper Bound} \\ \midrule
\ourmethod{} (CIFAR-10/100, PathMNIST)  & Sensitivity $\alpha$                & 0                  & 1                 \\ \midrule
\ourmethod{} (BBC News)  & Sensitivity $\alpha$                & 0                  & 3                 \\ \midrule
Label Smoothing   & Smoothing Parameter $\epsilon$                & $0$                  & $1$   \\ \midrule
Confidence Penalty & Penalty Parameter $\lambda$                & $0$                  & $10$                 \\ \midrule
Focal Loss & Focusing parameter $\gamma$                & $0$                  & $5$                 \\ \midrule
\midrule
CONEX + Squared Penalty & Penalty Parameter $\rho$ &$0$ &$10^{3}$ \\ \midrule
CONEX + AL & Penalty Parameter $\rho$ &$0$ &$10^{3}$ \\
 & Inverse Scaling Factor $\nu$ &$1$ &$10^{3}$ \\ \midrule
\end{tabular}
\end{adjustbox}
\end{center}
\end{table}

We note that we do not tune the learning rate, but fix it to a fixed value per experiment. The reason is that it would allow methods that tend to overfit to delay their overfitting beyond the considered range of tuning epochs instead of alleviating overfitting. For each experiment and method, the amount of epochs per tuning run is identical to the training horizon for the reported validation and test results.

\subsection{Hyperparameters for ImageNet} \label{appx:imagenet_params}

For reasons of computational requirements, we use all methods with their default parameters, as specified in the original works~(see~\cref{sec:experiments_setup} for reference). Specifically, we use the following values:

\begin{table}[h]
\begin{center}
\begin{adjustbox}{max width=\textwidth}
\begin{tabular}{@{}ll@{}}
\toprule
\textbf{Method} & \textbf{Parameter Setting} \\ \midrule
\ourmethod{} & Sensitivity $\alpha = 0.01$ \\ \midrule
Label Smoothing & Smoothing Parameter $\epsilon = 0.1$ (see~\citet{szegedy2016rethinking}) \\ \midrule
Focal Loss & Focusing Parameter $\gamma = 2$ (see~\citet{ross2017focal}) \\ \midrule
Confidence Penalty & \makecell[l]{Penalty Parameter $\lambda = 0.1$ (see~\citet{pereyra2017regularizing});\\
we use the reported value for CIFAR-10 because no ImageNet experiment is conducted} \\ \midrule
\end{tabular}
\end{adjustbox}
\end{center}
\end{table}

\newpage

\section{Additional Experiments}

\subsection{{Runtime Comparison}} \label{appx:runtime}

To quantify the computational overhead of \ourmethod{} relative to standard objectives, we benchmark the wall-clock time of a single forward loss evaluation for several classification losses in PyTorch. We consider a $K=10$ class problem and generate synthetic logits $\mathbf{z} \in \mathbb{R}^{64 \times 10}$ and labels $y \in \{0,\dots,9\}^{64}$, corresponding to a batch size of $64$. For each method we draw $N = 100{,}000$ independent batches of logits and labels from a fixed random seed and measure the time required for a single loss computation using \texttt{time.time()}. The results can be seen in the following table and show that \ourmethod{} even tends to be slightly faster to evaluate than other regularization techniques:

\begin{table}[h]
    \centering
    \begin{tabular}{lrr}
        \toprule
        Method & Avg. time [s] & Std [s] \\
        \midrule
        CE         & $1.9 \times 10^{-5}$ & $5 \times 10^{-6}$ \\
        label smoothing  & $7.4 \times 10^{-5}$ & $5 \times 10^{-6}$ \\
        confidence penalty    & $2.0 \times 10^{-4}$ & $6 \times 10^{-6}$ \\
        focal loss     & $1.07 \times 10^{-4}$ & $5 \times 10^{-6}$ \\
        PENEX      & $6.8 \times 10^{-5}$ & $7 \times 10^{-6}$ \\
        \bottomrule
    \end{tabular}
\end{table}

\subsection{All Validation Results} \label{appx:all_validation_curves}

\begin{figure}
    \centering
    \includegraphics[width=0.77\textwidth]{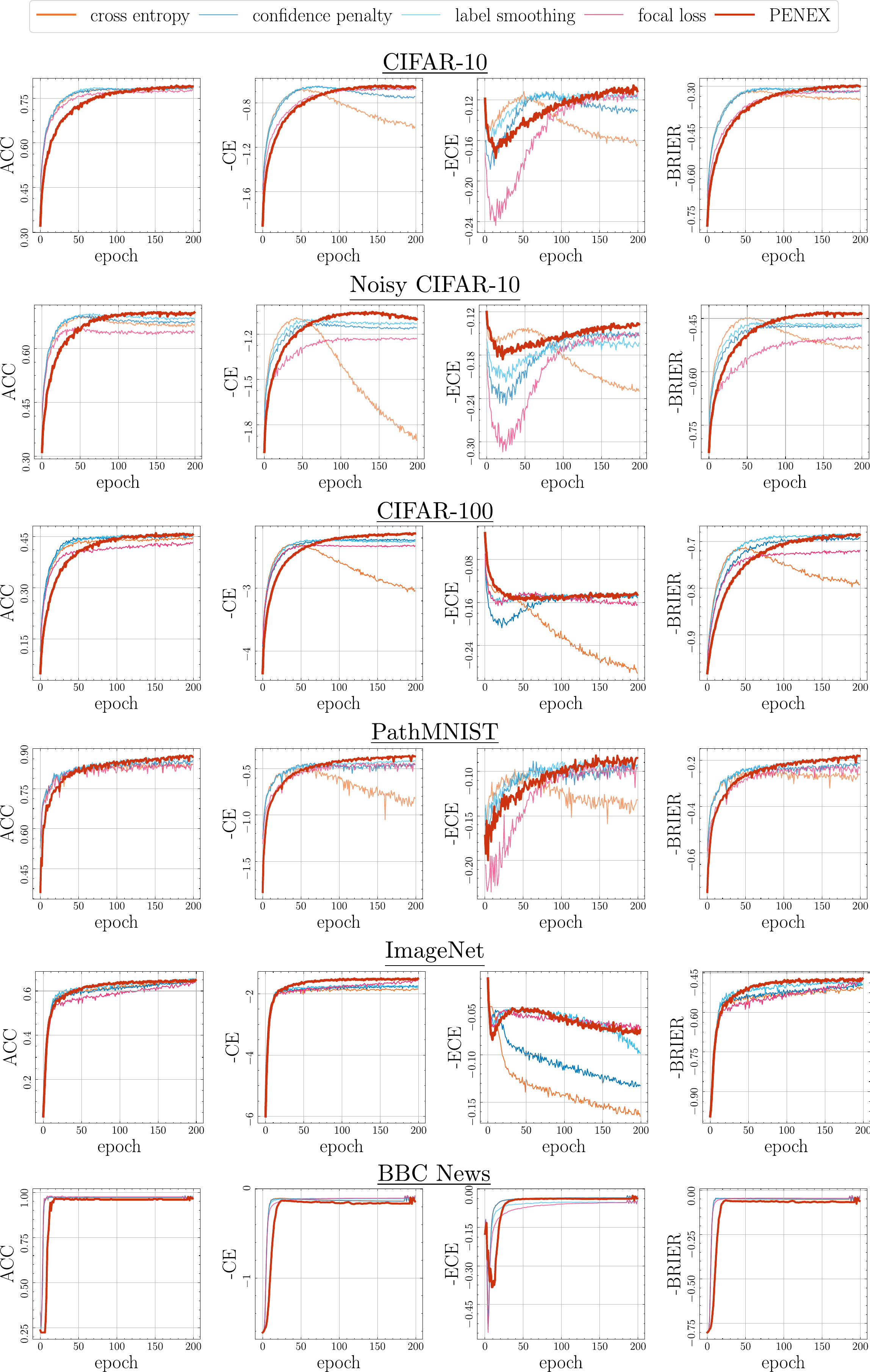}
    \caption{\textbf{All Validation Curves.} \underline{Larger means better}. Validation curves over $200$ epochs for all experiments, similar to~\cref{fig:metric_curves}.}
    \label{fig:metric_curves_all}
\end{figure}

The additional plots shown in~\cref{fig:metric_curves_all} support that~\ourmethod{} performs at least as good as all considered baseline regularizers and typically performs even better. The only clear exception is the ImageNet experiment, where we can observe difficulties in training stability, requiring a smaller learning rate (see~\cref{appx:training}) and leading to worse generalization as a consequence.

\subsection{Alternative Model Architecture} \label{appx:alternative_architecture_experiment}

To demonstrate that \ourmethod{} does not only work well for a single architecture, we generate test results for CIFAR-10 and CIFAR-100 with the ResNet-18 architecture~\citep{he2016deep} described in~\cref{appx:architectures}, with the training procedure described in~\cref{appx:training}. We use the default parameters for each method and additional weight decay of $5 \cdot 10^{-4}$ and momentum of $0.9$, for all methods. The results are provided in the following table:

\begin{table}[h]
    \centering
    \resizebox{0.99 \textwidth}{!}{
    \begin{tabular}{lccccc}
    \toprule
    Metric & CE & label smoothing & confidence penalty & focal loss & PENEX \\
    \midrule
    \multicolumn{6}{c}{\textbf{CIFAR-10}} \\
    ACC      & $0.947 \pm 0.002$ & $\mathbf{0.949 \pm 0.002}$ & $0.947 \pm 0.002$ & $0.945 \pm 0.002$ & $0.948 \pm 0.002$ \\
    $-\mathrm{ECE}$   & $-0.042 \pm 0.001$ & $-0.096 \pm 0.001$ & $-0.039 \pm 0.002$ & $-0.094 \pm 0.001$ & $\mathbf{-0.037 \pm 0.001}$ \\
    $-\mathrm{CE}$    & $-0.197 \pm 0.009$ & $-0.259 \pm 0.007$ & $\mathbf{-0.181 \pm 0.007}$ & $-0.223 \pm 0.004$ & $-0.193 \pm 0.008$ \\
    $-\mathrm{BRIER}$ & $-0.084 \pm 0.003$ & $-0.088 \pm 0.003$ & $\mathbf{-0.082 \pm 0.003}$ & $-0.084 \pm 0.002$ & $-0.083 \pm 0.003$ \\
    \midrule
    \multicolumn{6}{c}{\textbf{CIFAR-100}} \\
    ACC      & $0.571 \pm 0.005$ & $0.571 \pm 0.005$ & $0.549 \pm 0.005$ & $0.558 \pm 0.005$ & $\mathbf{0.601 \pm 0.005}$ \\
    $-\mathrm{ECE}$   & $-0.163 \pm 0.003$ & $\mathbf{-0.107 \pm 0.003}$ & $-0.157 \pm 0.004$ & $-0.112 \pm 0.003$ & $-0.112 \pm 0.002$ \\
    $-\mathrm{CE}$    & $-1.696 \pm 0.023$ & $-1.678 \pm 0.019$ & $-1.799 \pm 0.024$ & $-1.583 \pm 0.017$ & $\mathbf{-1.511 \pm 0.017}$ \\
    $-\mathrm{BRIER}$ & $-0.588 \pm 0.006$ & $-0.569 \pm 0.006$ & $-0.612 \pm 0.007$ & $-0.577 \pm 0.005$ & $\mathbf{-0.532 \pm 0.005}$ \\
    \bottomrule
    \end{tabular}
    }
\end{table}

\subsection{\revised{Additional Noisy Label Experiments}}\label{appx:additional_noise}
To evaluate whether the robustness of \ourmethod{} extends beyond the noise setting considered in the main experiments~\cref{sec:experiments_setup}, we conduct several additional experiments covering both synthetic and real-world label noise on CIFAR-10 with the main ConvNet implementation. 

\paragraph{Setup.}
First, we consider uniform label noise at two additional noise levels, $20\%$ and $40\%$. Second, we evaluate on real-world human annotation noise using the standardized CIFAR-10N benchmark introduced by~\citet{wei2022learning}. Unlike synthetically corrupted labels, these labels arise from actual human annotations and therefore provide a complementary evaluation under more realistic noise patterns. Finally, we consider class-asymmetric label noise on CIFAR-10. Following a fixed class-dependent corruption scheme, bird labels are changed to airplane, truck to automobile, cat to dog, dog to cat, and deer to horse, each with probability given by the noise rate, while the remaining classes are left unchanged. This setting introduces structured, class-dependent label corruption and therefore differs substantially from uniform random noise.

\paragraph{Results.}
    \begin{table}[h]
    \centering
    \resizebox{0.99\textwidth}{!}{
    \begin{tabular}{lccccc}
    \toprule
    Metric & CE & label smoothing & confidence penalty & focal loss & PENEX \\
\midrule
    \multicolumn{6}{c}{\textbf{20\% Uniform Label Noise}} \\
    ACC      & $0.682 \pm 0.004$ & $0.697 \pm 0.005$ & $0.685 \pm 0.004$ & $0.664 \pm 0.005$ & $\mathbf{0.702 \pm 0.004}$ \\
    $-\mathrm{ECE}$   & $-0.181 \pm 0.003$ & $-0.185 \pm 0.003$ & $\mathbf{-0.151 \pm 0.002}$ & $-0.169 \pm 0.003$ & $-0.160 \pm 0.003$ \\
    $-\mathrm{CE}$    & $-1.208 \pm 0.020$ & $-1.026 \pm 0.009$ & $-1.011 \pm 0.009$ & $-1.061 \pm 0.009$ & $\mathbf{-0.916 \pm 0.009}$ \\
    $-\mathrm{BRIER}$ & $-0.472 \pm 0.006$ & $-0.444 \pm 0.004$ & $-0.446 \pm 0.004$ & $-0.484 \pm 0.004$ & $\mathbf{-0.417 \pm 0.004}$ \\

    \midrule
    \multicolumn{6}{c}{\textbf{40\% Uniform Label Noise}} \\
    ACC      & $0.569 \pm 0.004$ & $0.590 \pm 0.005$ & $0.563 \pm 0.005$ & $0.565 \pm 0.005$ & $\mathbf{0.605 \pm 0.005}$ \\
    $-\mathrm{ECE}$   & $-0.197 \pm 0.003$ & $-0.171 \pm 0.003$ & $\mathbf{-0.155 \pm 0.003}$ & $-0.171 \pm 0.004$ & $-0.177 \pm 0.003$ \\
    $-\mathrm{CE}$    & $-1.529 \pm 0.019$ & $-1.338 \pm 0.009$ & $-1.389 \pm 0.011$ & $-1.371 \pm 0.009$ & $\mathbf{-1.245 \pm 0.009}$ \\
    $-\mathrm{BRIER}$ & $-0.606 \pm 0.006$ & $-0.573 \pm 0.004$ & $-0.593 \pm 0.004$ & $-0.605 \pm 0.004$ & $\mathbf{-0.552 \pm 0.004}$ \\

    \midrule
    \multicolumn{6}{c}{\textbf{Real-World Human Annotation Noise}} \\
    ACC      & $0.697 \pm 0.005$ & $\mathbf{0.717 \pm 0.004}$ & $0.703 \pm 0.004$ & $0.678 \pm 0.004$ & $\mathbf{0.717 \pm 0.005}$ \\
    $-\mathrm{ECE}$   & $-0.192 \pm 0.003$ & $-0.179 \pm 0.003$ & $-0.149 \pm 0.003$ & $-0.160 \pm 0.002$ & $\mathbf{-0.135 \pm 0.002}$ \\
    $-\mathrm{CE}$    & $-1.229 \pm 0.022$ & $-0.959 \pm 0.008$ & $-0.941 \pm 0.010$ & $-0.955 \pm 0.008$ & $\mathbf{-0.842 \pm 0.010}$ \\
    $-\mathrm{BRIER}$ & $-0.464 \pm 0.007$ & $-0.418 \pm 0.003$ & $-0.419 \pm 0.004$ & $-0.455 \pm 0.004$ & $\mathbf{-0.396 \pm 0.004}$ \\

    \midrule
    \multicolumn{6}{c}{\textbf{10\% Asymmetric Noise}} \\
    ACC      & $0.759 \pm 0.004$ & $\mathbf{0.767 \pm 0.004}$ & $0.759 \pm 0.004$ & $0.734 \pm 0.005$ & $0.765 \pm 0.004$ \\
    $-\mathrm{ECE}$   & $-0.172 \pm 0.003$ & $-0.185 \pm 0.003$ & $-0.156 \pm 0.002$ & $-0.142 \pm 0.003$ & $\mathbf{-0.120 \pm 0.002}$ \\
    $-\mathrm{CE}$    & $-1.087 \pm 0.023$ & $-0.813 \pm 0.007$ & $-0.792 \pm 0.008$ & $-0.780 \pm 0.009$ & $\mathbf{-0.721 \pm 0.012}$ \\
    $-\mathrm{BRIER}$ & $-0.382 \pm 0.006$ & $-0.351 \pm 0.004$ & $-0.353 \pm 0.004$ & $-0.380 \pm 0.004$ & $\mathbf{-0.329 \pm 0.004}$ \\

    \bottomrule
    \end{tabular}
    }
    \vspace{0.5em}
    \caption{Robustness to different forms of label noise on CIFAR-10. We report results for $20\%$ and $40\%$ uniform label noise, real-world human annotation noise from CIFAR-10N, and $10\%$ class-asymmetric label noise. \underline{Larger is better, best is bold}.}
    \label{tab:additional_noise}
\end{table}

The results, provided in~\cref{tab:additional_noise}, show that \ourmethod{} is consistently robust across all of the tested noise settings, outperforming most baselines on most metrics. Under both $20\%$ and $40\%$ uniform label noise, \ourmethod{} achieves the highest accuracy as well as the best cross-entropy and Brier scores, while remaining competitive in terms of ECE. The same trend extends to real-world human annotation noise, where \ourmethod{} attains the joint-highest accuracy and the best ECE, cross-entropy, and Brier scores. Finally, under asymmetric label noise, \ourmethod{} again obtains the best ECE, cross-entropy, and Brier scores, while its accuracy remains close to the best-performing baseline.

\subsection{\revised{Comparison with a Margin-Based Objective}}
\label{appx:margin_based}

\paragraph{Setup.}
To complement the comparisons against standard classification objectives, we additionally evaluate \ourmethod{} against mdNet~\citep{lyu2022improving}, a representative margin-based training objective designed to improve generalization by explicitly shaping the margin distribution. We evaluate all methods on CIFAR-10 using the same experimental protocol as in the main experiments. Cross-entropy (CE) is included as a reference baseline. The results are given in the following table:

\begin{table}[h]
    \centering
    \resizebox{0.7\textwidth}{!}{
    \begin{tabular}{lccc}
    \toprule
    Metric & CE & mdNet & PENEX \\
    \midrule
    ACC
    & $0.785 \pm 0.004$
    & $0.786 \pm 0.004$
    & $\mathbf{0.793 \pm 0.004}$ \\
    $-\mathrm{ECE}$
    & $-0.162 \pm 0.003$
    & $-0.419 \pm 0.003$
    & $\mathbf{-0.109 \pm 0.002}$ \\
    $-\mathrm{CE}$
    & $-1.004 \pm 0.024$
    & $-1.196 \pm 0.005$
    & $\mathbf{-0.646 \pm 0.012}$ \\
    $-\mathrm{BRIER}$
    & $-0.346 \pm 0.006$
    & $-0.525 \pm 0.002$
    & $\mathbf{-0.299 \pm 0.005}$ \\
    \bottomrule
    \end{tabular}
    }
    \vspace{0.5em}
    \caption{Comparison with the margin-based objective mdNet on CIFAR-10. Cross-entropy (CE) is included as a reference. \underline{Larger is better, best is bold}.}
    \label{tab:margin_based}
\end{table}

\paragraph{Results.}
As shown in \cref{tab:margin_based}, mdNet achieves accuracy comparable to standard cross-entropy training, but performs substantially worse on all probability-sensitive metrics when its logits are transformed using the standard softmax. In contrast, \ourmethod{} achieves the best result across all four metrics, including both accuracy and calibration. This comparison highlights an important distinction between the two approaches. While mdNet operates directly on the classification margin, \ourmethod{} additionally admits a closed-form link between its population-optimal logits and the complete class-probability vector, as established in \cref{prop:self_calibration}. Consequently, \ourmethod{} combines the generalization benefits associated with margin-based training with principled probability estimation.

\subsection{Sensitivity Parameter Analysis} \label{appx:parameter_analysis}

\begin{figure}
    \centering
    \includegraphics[width=0.99\textwidth]{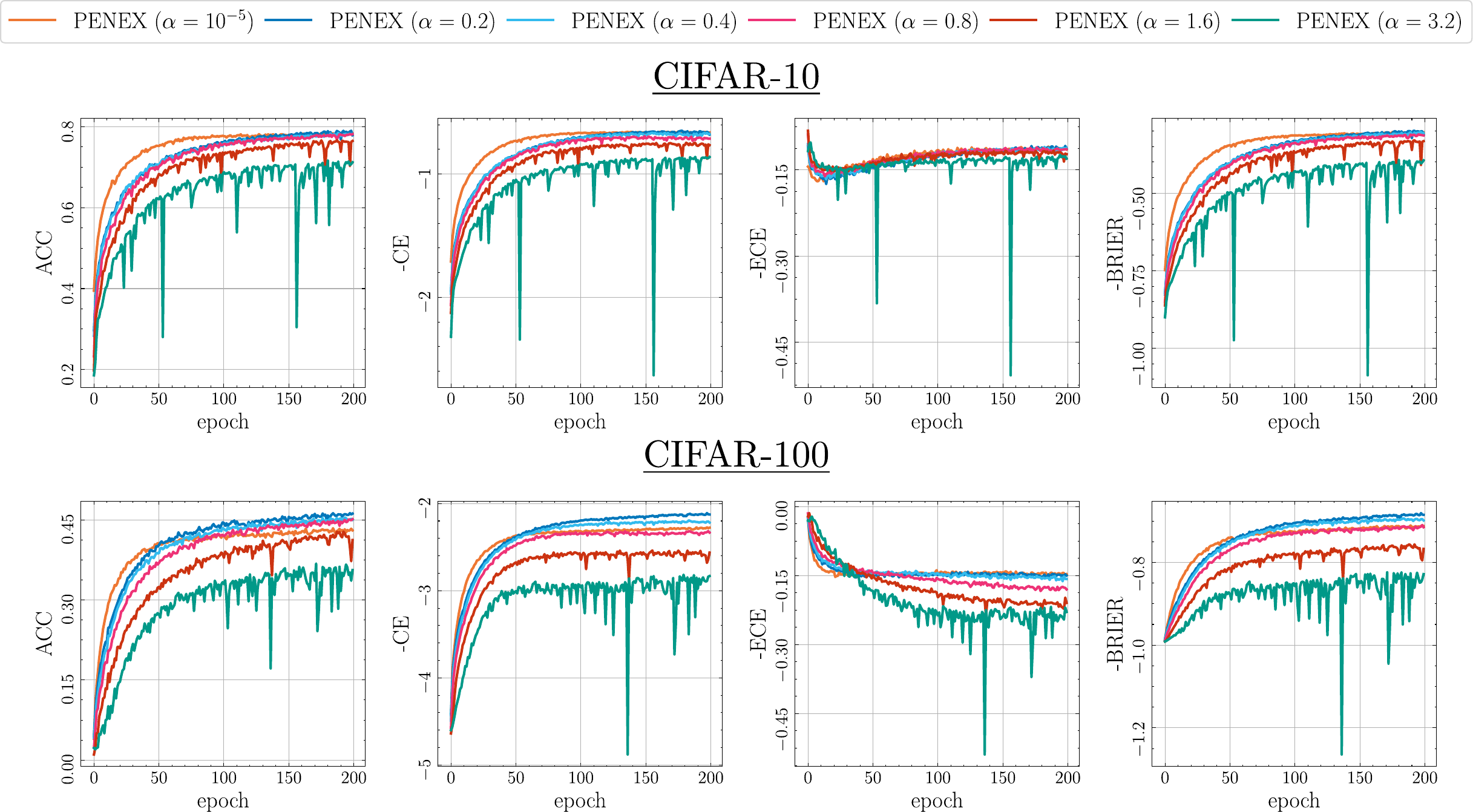}
    \caption{\textbf{Effects of the Sensitivity Parameter on Validation Curves.} Smaller $\alpha$ typically lead to faster convergence, but worse generalization toward the end of the training duration (as can be seen more clearly for CIFAR-100). In addition, for very large $\alpha = 3.2$, training curves tend to become less stable.}
    \label{fig:sensitivity_curves}
\end{figure}

The experimental results demonstrated in~\cref{fig:sensitivity_curves} show how the sensitivity parameter $\alpha$ of PENEX affects all metrics considered in the present work, for six different values $\alpha \in \{ 10^{-5}, 0.2, 0.4, 0.8, 1.6, 3.2 \} $. \cref{fig:sensitivity_curves} shows that larger values of $\alpha$ lead to slower improvements initially, but often better generalization toward the end of training. In addition, for very large $\alpha$, training curves become more unstable, likely due to steeper areas in the loss surface.

\subsection{\revised{Different Penalties}}
\label{appx:different_penalties}

\paragraph{Setup.}
To investigate whether the particular penalty formulation used by \ourmethod{} is important for its empirical performance, we compare the proposed SumExp penalty against two alternative formulations: a squared Euclidean penalty and a LogSumExp penalty. We evaluate all three variants using the same experimental protocol as in the main experiments. As throughout the paper, larger values are better for all reported metrics.

\begin{table}[h]
    \centering
    \resizebox{0.85\textwidth}{!}{
    \begin{tabular}{lccc}
    \toprule
    Metric & Squared Euclidean Penalty & LogSumExp Penalty & SumExp Penalty (PENEX) \\
    \midrule
    ACC
    & $0.765 \pm 0.004$
    & $0.757 \pm 0.004$
    & $\mathbf{0.793 \pm 0.004}$ \\
    $-\mathrm{ECE}$
    & $-0.664 \pm 0.004$
    & $-0.127 \pm 0.002$
    & $\mathbf{-0.109 \pm 0.002}$ \\
    $-\mathrm{CE}$
    & $-2.294 \pm 0.000$
    & $-0.709 \pm 0.010$
    & $\mathbf{-0.646 \pm 0.012}$ \\
    $-\mathrm{BRIER}$
    & $-0.898 \pm 0.000$
    & $-0.343 \pm 0.005$
    & $\mathbf{-0.299 \pm 0.005}$ \\
    \bottomrule
    \end{tabular}
    }
    \vspace{0.5em}
    \caption{Comparison of different penalty formulations. The SumExp penalty used by \ourmethod{} outperforms the squared Euclidean and LogSumExp alternatives across all reported metrics. \underline{Larger is better, best is bold}.}
    \label{tab:different_penalties}
\end{table}

\paragraph{Results.}
As shown in \cref{tab:different_penalties}, the alternative penalties yield viable classifiers, but neither matches the performance of the SumExp penalty used by \ourmethod{}. In particular, the LogSumExp penalty performs considerably better than the squared Euclidean alternative on the probability-sensitive metrics, yet remains consistently behind \ourmethod{} in accuracy, ECE, cross-entropy, and Brier score. The squared Euclidean penalty performs especially poorly in terms of calibration and probabilistic prediction quality. These results indicate that the empirical gains of \ourmethod{} are not merely a consequence of augmenting the exponential loss with an arbitrary logit penalty, but depend on the particular SumExp formulation. Moreover, the theoretical properties established for \ourmethod{}, including its self-calibration and margin-tail results, do not readily extend to these alternative penalties.

\subsection{\revised{Fixed Penalty Parameter}}
\label{appx:fixed_rho}

\paragraph{Setup.}
In the main experiments, we estimate $\rho$ automatically rather than treating it as an additional hyperparameter~(see~\cref{alg:penex_training}). To assess the cost of this design choice, we compare the default version of \ourmethod{} against a variant in which $\rho$ is fixed and tuned using the validation set. This comparison isolates whether explicitly tuning $\rho$ can provide additional performance beyond the automatic estimator used throughout the main experiments.

\begin{table}[h]
    \centering
    \resizebox{0.55\textwidth}{!}{
    \begin{tabular}{lcc}
    \toprule
    Metric & Default PENEX & Tuned fixed $\rho$ \\
    \midrule
    ACC
    & $0.786 \pm 0.004$
    & $\mathbf{0.797 \pm 0.004}$ \\
    $-\mathrm{ECE}$
    & $-0.114 \pm 0.002$
    & $\mathbf{-0.108 \pm 0.002}$ \\
    $-\mathrm{CE}$
    & $-0.676 \pm 0.014$
    & $\mathbf{-0.623 \pm 0.013}$ \\
    $-\mathrm{BRIER}$
    & $-0.304 \pm 0.005$
    & $\mathbf{-0.288 \pm 0.005}$ \\
    \bottomrule
    \end{tabular}
    }
    \vspace{0.5em}
    \caption{Comparison between the automatic estimator for $\rho$ used by default in \ourmethod{} and a validation-tuned fixed $\rho$. \underline{Larger is better, best is bold}.}
    \label{tab:fixed_rho}
\end{table}

\paragraph{Results.}
As shown in \cref{tab:fixed_rho}, explicitly tuning a fixed value of $\rho$ yields better performance across all four metrics. This improvement, however, comes at the cost of introducing a second loss hyperparameter that must be selected using validation data. Since the main baselines considered in our experiments have only one tuned loss parameter, we use the automatic estimator for $\rho$ in the main comparison and tune only the remaining loss parameter. The automatic estimator therefore reduces the tuning burden while retaining competitive performance, whereas validation-tuning $\rho$ provides an additional performance gain when the extra tuning cost is acceptable.

\subsection{\revised{Ablating EMA and Clipping Parameters}}
\label{appx:ema_clip_ablation}

\paragraph{Setup.}
We additionally study the sensitivity of \ourmethod{} to two implementation choices used by the automatic estimator for penalty parameter $\rho$: clipping and exponential moving averaging (EMA). We compare the default configuration against a variant without clipping and a variant using $\beta = 0.01$. This ablation tests whether the performance of \ourmethod{} depends sensitively on either the clipping bounds or the EMA factor.

\begin{table}[h]
    \centering
    \resizebox{0.7\textwidth}{!}{
    \begin{tabular}{lccc}
    \toprule
    Metric & Default PENEX & No clipping & $\beta = 0.01$ \\
    \midrule
    ACC
    & $0.786 \pm 0.004$
    & $0.786 \pm 0.004$
    & $0.785 \pm 0.004$ \\
    $-\mathrm{ECE}$
    & $-0.114 \pm 0.002$
    & $-0.114 \pm 0.002$
    & $-0.113 \pm 0.002$ \\
    $-\mathrm{CE}$
    & $-0.676 \pm 0.014$
    & $-0.676 \pm 0.014$
    & $-0.674 \pm 0.013$ \\
    $-\mathrm{BRIER}$
    & $-0.304 \pm 0.005$
    & $-0.304 \pm 0.005$
    & $-0.306 \pm 0.005$ \\
    \bottomrule
    \end{tabular}
    }
    \vspace{0.5em}
    \caption{Ablation of clipping and the EMA factor used by the automatic $\rho$ estimator. Performance remains nearly unchanged when clipping is removed or when the EMA factor is set to $\beta = 0.01$.}
    \label{tab:ema_clip_ablation}
\end{table}

\paragraph{Results.}
The three configurations in \cref{tab:ema_clip_ablation} perform almost identically across all reported metrics. Removing clipping leaves the reported results unchanged, while changing the EMA factor to $\beta = 0.01$ produces only negligible differences. These results indicate that neither the clipping bounds nor the EMA factor constitute sensitive hyperparameters of \ourmethod{}. In particular, the stability of the automatic $\rho$ estimator does not appear to rely on careful tuning of these implementation choices.

\newpage
\section*{NeurIPS Paper Checklist}

\begin{enumerate}

\item {\bf Claims}
    \item[] Question: Do the main claims made in the abstract and introduction accurately reflect the paper's contributions and scope?
    \item[] Answer: \answerYes{} %
    \item[] Justification: See abstract and~\cref{sec:introduction}.
    \item[] Guidelines:
    \begin{itemize}
        \item The answer \answerNA{} means that the abstract and introduction do not include the claims made in the paper.
        \item The abstract and/or introduction should clearly state the claims made, including the contributions made in the paper and important assumptions and limitations. A \answerNo{} or \answerNA{} answer to this question will not be perceived well by the reviewers. 
        \item The claims made should match theoretical and experimental results, and reflect how much the results can be expected to generalize to other settings. 
        \item It is fine to include aspirational goals as motivation as long as it is clear that these goals are not attained by the paper. 
    \end{itemize}
    
\item {\bf Limitations}
    \item[] Question: Does the paper discuss the limitations of the work performed by the authors?
    \item[] Answer: \answerYes{} %
    \item[] Justification: A discussion \& limitations section~(\cref{sec:discussion}) is included.
    \item[] Guidelines:
    \begin{itemize}
        \item The answer \answerNA{} means that the paper has no limitation while the answer \answerNo{} means that the paper has limitations, but those are not discussed in the paper. 
        \item The authors are encouraged to create a separate ``Limitations'' section in their paper.
        \item The paper should point out any strong assumptions and how robust the results are to violations of these assumptions (e.g., independence assumptions, noiseless settings, model well-specification, asymptotic approximations only holding locally). The authors should reflect on how these assumptions might be violated in practice and what the implications would be.
        \item The authors should reflect on the scope of the claims made, e.g., if the approach was only tested on a few datasets or with a few runs. In general, empirical results often depend on implicit assumptions, which should be articulated.
        \item The authors should reflect on the factors that influence the performance of the approach. For example, a facial recognition algorithm may perform poorly when image resolution is low or images are taken in low lighting. Or a speech-to-text system might not be used reliably to provide closed captions for online lectures because it fails to handle technical jargon.
        \item The authors should discuss the computational efficiency of the proposed algorithms and how they scale with dataset size.
        \item If applicable, the authors should discuss possible limitations of their approach to address problems of privacy and fairness.
        \item While the authors might fear that complete honesty about limitations might be used by reviewers as grounds for rejection, a worse outcome might be that reviewers discover limitations that aren't acknowledged in the paper. The authors should use their best judgment and recognize that individual actions in favor of transparency play an important role in developing norms that preserve the integrity of the community. Reviewers will be specifically instructed to not penalize honesty concerning limitations.
    \end{itemize}

    \item {\bf Theory assumptions and proofs}
    \item[] Question: For each theoretical result, does the paper provide the full set of assumptions and a complete (and correct) proof?
    \item[] Answer: \answerYes{} %
    \item[] Justification: A proof for each theoretical result is supplied in~\cref{appx:proofs}.
    \item[] Guidelines:
    \begin{itemize}
        \item The answer \answerNA{} means that the paper does not include theoretical results. 
        \item All the theorems, formulas, and proofs in the paper should be numbered and cross-referenced.
        \item All assumptions should be clearly stated or referenced in the statement of any theorems.
        \item The proofs can either appear in the main paper or the supplemental material, but if they appear in the supplemental material, the authors are encouraged to provide a short proof sketch to provide intuition. 
        \item Inversely, any informal proof provided in the core of the paper should be complemented by formal proofs provided in appendix or supplemental material.
        \item Theorems and Lemmas that the proof relies upon should be properly referenced. 
    \end{itemize}
    
    \item {\bf Experimental result reproducibility}
    \item[] Question: Does the paper fully disclose all the information needed to reproduce the main experimental results of the paper to the extent that it affects the main claims and/or conclusions of the paper (regardless of whether the code and data are provided or not)?
    \item[] Answer: \answerYes{} %
    \item[] Justification: In addition to source code, detailed descriptions for optimal reproducibility are supplied.
    \item[] Guidelines:
    \begin{itemize}
        \item The answer \answerNA{} means that the paper does not include experiments.
        \item If the paper includes experiments, a \answerNo{} answer to this question will not be perceived well by the reviewers: Making the paper reproducible is important, regardless of whether the code and data are provided or not.
        \item If the contribution is a dataset and\slash or model, the authors should describe the steps taken to make their results reproducible or verifiable. 
        \item Depending on the contribution, reproducibility can be accomplished in various ways. For example, if the contribution is a novel architecture, describing the architecture fully might suffice, or if the contribution is a specific model and empirical evaluation, it may be necessary to either make it possible for others to replicate the model with the same dataset, or provide access to the model. In general. releasing code and data is often one good way to accomplish this, but reproducibility can also be provided via detailed instructions for how to replicate the results, access to a hosted model (e.g., in the case of a large language model), releasing of a model checkpoint, or other means that are appropriate to the research performed.
        \item While NeurIPS does not require releasing code, the conference does require all submissions to provide some reasonable avenue for reproducibility, which may depend on the nature of the contribution. For example
        \begin{enumerate}
            \item If the contribution is primarily a new algorithm, the paper should make it clear how to reproduce that algorithm.
            \item If the contribution is primarily a new model architecture, the paper should describe the architecture clearly and fully.
            \item If the contribution is a new model (e.g., a large language model), then there should either be a way to access this model for reproducing the results or a way to reproduce the model (e.g., with an open-source dataset or instructions for how to construct the dataset).
            \item We recognize that reproducibility may be tricky in some cases, in which case authors are welcome to describe the particular way they provide for reproducibility. In the case of closed-source models, it may be that access to the model is limited in some way (e.g., to registered users), but it should be possible for other researchers to have some path to reproducing or verifying the results.
        \end{enumerate}
    \end{itemize}
    
\item {\bf Open access to data and code}
    \item[] Question: Does the paper provide open access to the data and code, with sufficient instructions to faithfully reproduce the main experimental results, as described in supplemental material?
    \item[] Answer: \answerYes{} %
    \item[] Justification: Source code is supplied. All used datasets are publicly available and referenced.
    \item[] Guidelines:
    \begin{itemize}
        \item The answer \answerNA{} means that paper does not include experiments requiring code.
        \item Please see the NeurIPS code and data submission guidelines (\url{https://neurips.cc/public/guides/CodeSubmissionPolicy}) for more details.
        \item While we encourage the release of code and data, we understand that this might not be possible, so \answerNo{} is an acceptable answer. Papers cannot be rejected simply for not including code, unless this is central to the contribution (e.g., for a new open-source benchmark).
        \item The instructions should contain the exact command and environment needed to run to reproduce the results. See the NeurIPS code and data submission guidelines (\url{https://neurips.cc/public/guides/CodeSubmissionPolicy}) for more details.
        \item The authors should provide instructions on data access and preparation, including how to access the raw data, preprocessed data, intermediate data, and generated data, etc.
        \item The authors should provide scripts to reproduce all experimental results for the new proposed method and baselines. If only a subset of experiments are reproducible, they should state which ones are omitted from the script and why.
        \item At submission time, to preserve anonymity, the authors should release anonymized versions (if applicable).
        \item Providing as much information as possible in supplemental material (appended to the paper) is recommended, but including URLs to data and code is permitted.
    \end{itemize}
    
\item {\bf Experimental setting/details}
    \item[] Question: Does the paper specify all the training and test details (e.g., data splits, hyperparameters, how they were chosen, type of optimizer) necessary to understand the results?
    \item[] Answer: \answerYes{} %
    \item[] Justification: A detailed description of all hyperparameters is supplied in~\cref{appx:experimental_details}.
    \item[] Guidelines:
    \begin{itemize}
        \item The answer \answerNA{} means that the paper does not include experiments.
        \item The experimental setting should be presented in the core of the paper to a level of detail that is necessary to appreciate the results and make sense of them.
        \item The full details can be provided either with the code, in appendix, or as supplemental material.
    \end{itemize}

\item {\bf Experiment statistical significance}
    \item[] Question: Does the paper report error bars suitably and correctly defined or other appropriate information about the statistical significance of the experiments?
    \item[] Answer: \answerYes{} %
    \item[] Justification: Standard deviations (and a description of how they are estimated) are supplied for all experimental results.
    \item[] Guidelines:
    \begin{itemize}
        \item The answer \answerNA{} means that the paper does not include experiments.
        \item The authors should answer \answerYes{} if the results are accompanied by error bars, confidence intervals, or statistical significance tests, at least for the experiments that support the main claims of the paper.
        \item The factors of variability that the error bars are capturing should be clearly stated (for example, train/test split, initialization, random drawing of some parameter, or overall run with given experimental conditions).
        \item The method for calculating the error bars should be explained (closed form formula, call to a library function, bootstrap, etc.)
        \item The assumptions made should be given (e.g., Normally distributed errors).
        \item It should be clear whether the error bar is the standard deviation or the standard error of the mean.
        \item It is OK to report 1-sigma error bars, but one should state it. The authors should preferably report a 2-sigma error bar than state that they have a 96\% CI, if the hypothesis of Normality of errors is not verified.
        \item For asymmetric distributions, the authors should be careful not to show in tables or figures symmetric error bars that would yield results that are out of range (e.g., negative error rates).
        \item If error bars are reported in tables or plots, the authors should explain in the text how they were calculated and reference the corresponding figures or tables in the text.
    \end{itemize}

\item {\bf Experiments compute resources}
    \item[] Question: For each experiment, does the paper provide sufficient information on the computer resources (type of compute workers, memory, time of execution) needed to reproduce the experiments?
    \item[] Answer: \answerYes{} %
    \item[] Justification: Information about computational resources is available in~\cref{appx:computational_ressources}.
    \item[] Guidelines:
    \begin{itemize}
        \item The answer \answerNA{} means that the paper does not include experiments.
        \item The paper should indicate the type of compute workers CPU or GPU, internal cluster, or cloud provider, including relevant memory and storage.
        \item The paper should provide the amount of compute required for each of the individual experimental runs as well as estimate the total compute. 
        \item The paper should disclose whether the full research project required more compute than the experiments reported in the paper (e.g., preliminary or failed experiments that didn't make it into the paper). 
    \end{itemize}

\item {\bf Code of ethics}
    \item[] Question: Does the research conducted in the paper conform, in every respect, with the NeurIPS Code of Ethics \url{https://neurips.cc/public/EthicsGuidelines}?
    \item[] Answer: \answerYes{} %
    \item[] Justification: There are no conflicts with the NeurIPS Code of Ethics.
    \item[] Guidelines:
    \begin{itemize}
        \item The answer \answerNA{} means that the authors have not reviewed the NeurIPS Code of Ethics.
        \item If the authors answer \answerNo, they should explain the special circumstances that require a deviation from the Code of Ethics.
        \item The authors should make sure to preserve anonymity (e.g., if there is a special consideration due to laws or regulations in their jurisdiction).
    \end{itemize}

\item {\bf Broader impacts}
    \item[] Question: Does the paper discuss both potential positive societal impacts and negative societal impacts of the work performed?
    \item[] Answer: \answerNA{} %
    \item[] Justification: The contribution, which is a new loss function and its properties, is quite general and does not serve a specific purpose that would render such a discussion necessary. 
    \item[] Guidelines:
    \begin{itemize}
        \item The answer \answerNA{} means that there is no societal impact of the work performed.
        \item If the authors answer \answerNA{} or \answerNo, they should explain why their work has no societal impact or why the paper does not address societal impact.
        \item Examples of negative societal impacts include potential malicious or unintended uses (e.g., disinformation, generating fake profiles, surveillance), fairness considerations (e.g., deployment of technologies that could make decisions that unfairly impact specific groups), privacy considerations, and security considerations.
        \item The conference expects that many papers will be foundational research and not tied to particular applications, let alone deployments. However, if there is a direct path to any negative applications, the authors should point it out. For example, it is legitimate to point out that an improvement in the quality of generative models could be used to generate Deepfakes for disinformation. On the other hand, it is not needed to point out that a generic algorithm for optimizing neural networks could enable people to train models that generate Deepfakes faster.
        \item The authors should consider possible harms that could arise when the technology is being used as intended and functioning correctly, harms that could arise when the technology is being used as intended but gives incorrect results, and harms following from (intentional or unintentional) misuse of the technology.
        \item If there are negative societal impacts, the authors could also discuss possible mitigation strategies (e.g., gated release of models, providing defenses in addition to attacks, mechanisms for monitoring misuse, mechanisms to monitor how a system learns from feedback over time, improving the efficiency and accessibility of ML).
    \end{itemize}   

\item {\bf Safeguards}
    \item[] Question: Does the paper describe safeguards that have been put in place for responsible release of data or models that have a high risk for misuse (e.g., pre-trained language models, image generators, or scraped datasets)?
    \item[] Answer: \answerNA{} %
    \item[] Justification: There is no high risk of misuse.
    \item[] Guidelines:
    \begin{itemize}
        \item The answer \answerNA{} means that the paper poses no such risks.
        \item Released models that have a high risk for misuse or dual-use should be released with necessary safeguards to allow for controlled use of the model, for example by requiring that users adhere to usage guidelines or restrictions to access the model or implementing safety filters. 
        \item Datasets that have been scraped from the Internet could pose safety risks. The authors should describe how they avoided releasing unsafe images.
        \item We recognize that providing effective safeguards is challenging, and many papers do not require this, but we encourage authors to take this into account and make a best faith effort.
    \end{itemize}

\item {\bf Licenses for existing assets}
    \item[] Question: Are the creators or original owners of assets (e.g., code, data, models), used in the paper, properly credited and are the license and terms of use explicitly mentioned and properly respected?
    \item[] Answer: \answerYes{} %
    \item[] Justification: All sources of both data and models are credited, according to their license and terms (see~\cref{sec:experiments}). Licenses are explicitly listed in~\cref{appx:licenses}.
    \item[] Guidelines:
    \begin{itemize}
        \item The answer \answerNA{} means that the paper does not use existing assets.
        \item The authors should cite the original paper that produced the code package or dataset.
        \item The authors should state which version of the asset is used and, if possible, include a URL.
        \item The name of the license (e.g., CC-BY 4.0) should be included for each asset.
        \item For scraped data from a particular source (e.g., website), the copyright and terms of service of that source should be provided.
        \item If assets are released, the license, copyright information, and terms of use in the package should be provided. For popular datasets, \url{paperswithcode.com/datasets} has curated licenses for some datasets. Their licensing guide can help determine the license of a dataset.
        \item For existing datasets that are re-packaged, both the original license and the license of the derived asset (if it has changed) should be provided.
        \item If this information is not available online, the authors are encouraged to reach out to the asset's creators.
    \end{itemize}

\item {\bf New assets}
    \item[] Question: Are new assets introduced in the paper well documented and is the documentation provided alongside the assets?
    \item[] Answer: \answerNA{} %
    \item[] Justification: No new assets are released.
    \item[] Guidelines:
    \begin{itemize}
        \item The answer \answerNA{} means that the paper does not release new assets.
        \item Researchers should communicate the details of the dataset\slash code\slash model as part of their submissions via structured templates. This includes details about training, license, limitations, etc. 
        \item The paper should discuss whether and how consent was obtained from people whose asset is used.
        \item At submission time, remember to anonymize your assets (if applicable). You can either create an anonymized URL or include an anonymized zip file.
    \end{itemize}
    
\item {\bf Crowdsourcing and research with human subjects}
    \item[] Question: For crowdsourcing experiments and research with human subjects, does the paper include the full text of instructions given to participants and screenshots, if applicable, as well as details about compensation (if any)? 
    \item[] Answer: \answerNA{} %
    \item[] Justification: No crowdsourcing and no research with human subjects is involved.
    \item[] Guidelines:
    \begin{itemize}
        \item The answer \answerNA{} means that the paper does not involve crowdsourcing nor research with human subjects.
        \item Including this information in the supplemental material is fine, but if the main contribution of the paper involves human subjects, then as much detail as possible should be included in the main paper. 
        \item According to the NeurIPS Code of Ethics, workers involved in data collection, curation, or other labor should be paid at least the minimum wage in the country of the data collector. 
    \end{itemize}

\item {\bf Institutional review board (IRB) approvals or equivalent for research with human subjects}
    \item[] Question: Does the paper describe potential risks incurred by study participants, whether such risks were disclosed to the subjects, and whether Institutional Review Board (IRB) approvals (or an equivalent approval/review based on the requirements of your country or institution) were obtained?
    \item[] Answer: \answerNA{} %
    \item[] Justification: No crowdsourcing and no research with human subjects is involved.
    \item[] Guidelines:
    \begin{itemize}
        \item The answer \answerNA{} means that the paper does not involve crowdsourcing nor research with human subjects.
        \item Depending on the country in which research is conducted, IRB approval (or equivalent) may be required for any human subjects research. If you obtained IRB approval, you should clearly state this in the paper. 
        \item We recognize that the procedures for this may vary significantly between institutions and locations, and we expect authors to adhere to the NeurIPS Code of Ethics and the guidelines for their institution. 
        \item For initial submissions, do not include any information that would break anonymity (if applicable), such as the institution conducting the review.
    \end{itemize}
    
\item {\bf Declaration of LLM usage}
    \item[] Question: Does the paper describe the usage of LLMs if it is an important, original, or non-standard component of the core methods in this research? Note that if the LLM is used only for writing, editing, or formatting purposes and does \emph{not} impact the core methodology, scientific rigor, or originality of the research, declaration is not required.
    \item[] Answer: \answerNA{} %
    \item[] Justification: The core method development in this research does not involve LLMs.
    \item[] Guidelines:
    \begin{itemize}
        \item The answer \answerNA{} means that the core method development in this research does not involve LLMs as any important, original, or non-standard components.
        \item Please refer to our LLM policy in the NeurIPS handbook for what should or should not be described.
    \end{itemize}

\end{enumerate}

\end{document}